\pdfoutput=1

\documentclass[11pt]{article}

\usepackage{ACL2023}

\usepackage{times}
\usepackage{latexsym}

\usepackage[T1]{fontenc}

\usepackage[utf8]{inputenc}

\usepackage{microtype}

\usepackage{inconsolata}

\usepackage{graphicx}
\usepackage{multirow}
\usepackage[]{xcolor}
\title{Reference Matters: Benchmarking Factual Error Correction for Dialogue Summarization with Fine-grained Evaluation Framework}

\author{Mingqi Gao$^{1,2,3}$, Xiaojun Wan$^{1,2,3}$, Jia Su$^{4}$, Zhefeng Wang$^{4}$, Baoxing Huai$^{4}$\\
         $^{1}$Wangxuan Institute of Computer Technology, Peking University\\
         $^{2}$Center for Data Science, Peking University\\
         $^{3}$The MOE Key Laboratory of Computational Linguistics, Peking University\\
         $^{4}$Huawei Cloud\\
         \{gaomingqi,wanxiaojun\}@pku.edu.cn\\ \{sujia3,wangzhefeng,huaibaoxing\}@huawei.com}

\begin{document}
\maketitle
\begin{abstract}
Factuality is important to dialogue summarization. Factual error correction (FEC) of model-generated summaries is one way to improve factuality. Current FEC evaluation that relies on factuality metrics is not reliable and detailed enough. To address this problem, we are the first to manually annotate a FEC dataset for dialogue summarization containing 4000 items and propose FERRANTI, a fine-grained evaluation framework based on reference correction that automatically evaluates the performance of FEC models on different error categories. Using this evaluation framework, we conduct sufficient experiments with FEC approaches under a variety of settings and find the best training modes and significant differences in the performance of the existing approaches on different factual error categories. \footnote{Code and data will be available at \url{https://github.com/kite99520/DialSummFactCorr}}
\end{abstract}

\section{Introduction}
\label{sec:intro}
Factuality (also known as factual consistency, faithfulness) is a crucial dimension in evaluating summary quality. The summaries generated by current summarization models, and even some reference summaries, still have much room for improvement in factuality (\citealp{maynez-etal-2020-faithfulness}; \citealp{fabbri-etal-2021-summeval}; \citealp{pagnoni-etal-2021-understanding}). 
Dialogue summarization, a recently popular subfield of text summarization, has more challenging factual issues involved (\citealp{wang2022analyzing}; \citealp{gao-wan-2022-dialsummeval}).

The prior approaches to enhance the factuality of summaries can be broadly classified into two categories: one is to introduce factuality-related objectives in training or inference process to make the summarization models more faithful, which is a direct generation of factually better summaries (\citealp{falke-etal-2019-ranking}; \citealp{liu-chen-2021-controllable}; \citealp{wan-bansal-2022-factpegasus}; \citealp{tang-etal-2022-confit}; \citealp{liu-etal-2021-coreference}); the other is to design a factual error correction (FEC) model independent of the summarization models, which takes the source document and the summary to be corrected as input and outputs a corrected summary (\citealp{cao-etal-2020-factual}; \citealp{dong-etal-2020-multi}; \citealp{zhu-etal-2021-enhancing}; \citealp{chen-etal-2021-improving}; \citealp{Fabbri2022ImprovingFC}; \citealp{Balachandran2022CorrectingDF}). There are a number of studies on news summarization that can fall into both categories. To the best of our knowledge, there has been no work on factual error correction for dialogue summarization. Considering the importance of factual issues in dialogue summarization, we would like to try to correct factual errors in dialogue summaries.

However, after carefully examining and considering the motivations and practices of previous FEC studies, we argue that there are flaws in the way FEC models are evaluated, which may have diverted the FEC for summarization from its original purpose. Previous studies evaluate the effectiveness of FEC models mainly by judging whether the scores of factuality metrics (e.g. FactCC \citep{kryscinski-etal-2020-evaluating}) of the corrected summaries increase compared to the original summaries. First, this evaluation mechanism is so vague that it is difficult to evaluate the effectiveness of factual error correction accurately: we neither know which parts of the original summary have factual errors nor whether the corrected summary addresses them as expected.
Second, this evaluation mechanism also blurs the line between FEC for summarization and the direct generation of factually better summaries: the factual error correction model can ignore the content of the original summary and directly generate a different but more factually correct summary.

We argue that it is necessary to introduce manually annotated reference correction to address the above issues. Factual error correction for summarization has its basic requirement: to correct factual errors in the original summary by as few substitution, insertion, and deletion operations as possible to obtain a fluent and non-redundant summary. This can be reflected in the manual annotation. The introduction of reference correction, on the one hand, provides more valuable data for the training of FEC models compared to pseudo data; on the other hand, and more importantly, it creates the condition for a more comprehensive and accurate evaluation of the performance of FEC models. We construct an evaluation framework that can assess the performance of FEC models on different factual error categories based on manually annotated references. Using this framework, we are able to comprehensively evaluate and analyze the performance of various FEC methods on dialogue summarization. Our work has the following three main contributions:

\begin{itemize}
  \item [1)]
We collect the outputs of four common models on two dialogue summarization datasets and are the first to correct the factual errors in them manually. The dataset containing 4000 data items will be released to facilitate further research.
  \item [2)]
We propose FERRANTI, a fine-grained evaluation framework based on reference correction that provides a comprehensive assessment of the performance of FEC models on different categories of factual errors. 
  \item [3)]
Based on the above dataset and evaluation framework, we conduct a comprehensive evaluation and analysis of the performance of multiple FEC methods for dialogue summarization under different settings to illustrate the role of manually annotated data and the weaknesses of current models.
\end{itemize}

\section{Related Work}

\subsection{Dialogue Summarization Models} 
As datasets such as SAMSum \citep{gliwa-etal-2019-samsum} were proposed, many models designed for dialogue summarization sprang up. Many of them build on generic pre-trained generative models such as BART \citep{lewis-etal-2020-bart}, incorporating dialogue structure information such as multiple views \citep{chen-yang-2020-multi}, summary sketch \citep{wu-etal-2021-controllable}, argument mining \citep{fabbri-etal-2021-convosumm}, personal named entity \citep{liu-chen-2021-controllable}, and discourse relations \citep{chen-yang-2021-structure}. 
The summaries generated by these systems contain factual errors. They are what the FEC model needs to correct.

\subsection{FEC for Summarization} 
\citet{cao-etal-2020-factual} and \citet{dong-etal-2020-multi} can be considered as the first work on FEC for text summarization. \citet{cao-etal-2020-factual} apply data augmentation methods to transform the reference summary, obtain pseudo data to fine-tune the pre-trained model, and generate the corrected summary directly. In contrast,  \citet{dong-etal-2020-multi} use a more conservative strategy: masking the entities in summary and training a QA model to select span as the answer from the source document. \citet{Balachandran2022CorrectingDF} follow the idea of \citet{cao-etal-2020-factual} and generate harder pseudo data through infilling language models. A similar approach based on data augmentation is \citet{zhu-etal-2021-enhancing}, which makes use of the knowledge graph extracted from the source document. \citet{chen-etal-2021-improving} 
replace named entities and numbers in the summary 
to generate candidates, from which the best one is selected as the corrected summary. In addition, \citet{Fabbri2022ImprovingFC} train the model using sentence-compressed data and remove hallucinated entities from the summary.
We will test some of these methods on real annotated data of dialogue summarization.

\subsection{Factuality Evaluation for Summarization}
There are two main types of metrics widely used to evaluate the factuality of summaries. A class of metrics based on natural language inference, which formulate factuality as the result or confidence of binary classification, such as FactCC \citep{kryscinski-etal-2020-evaluating}, DAE (\citealp{goyal-durrett-2020-evaluating}; \citealp{goyal-durrett-2021-annotating}), and SUMMAC \citep{laban-etal-2022-summac}.  The other class is QA-based metrics, which usually contain a module for question generation and a module for question answering, with different implementation details, such as FEQA \citep{durmus-etal-2020-feqa}, SummaQA \citep{scialom-etal-2019-answers}, QuestEval \citep{scialom-etal-2021-questeval}, and QAFactEval \citep{fabbri-etal-2022-qafacteval}. 
Besides, BARTScore \citep{yuan2021bartscore} is also used to assess factuality. Many of them are used to evaluate the effectiveness of FEC models for summarization. 

\subsection{Evaluation for Post-editing and Correction}
Evidence-based factual error correction is to correct the factual errors in a claim with evidence texts from trustworthy knowledge bases (\citealp{thorne-vlachos-2021-evidence}; \citealp{Shah_Schuster_Barzilay_2020}; \citealp{Chen2022ConvergeTT}). Reference-based evaluation metrics SARI \citep{xu-etal-2016-optimizing} and ROUGE correlate highly with human judgments on evidence-based FEC \citep{thorne-vlachos-2021-evidence}. Automatic post-editing (APE) of machine translation and grammar error correction (GEC) also mainly use reference-based metrics \citep{chollampatt-etal-2020-automatic}. For APE, they are BLEU, TER \citep{snover2006ter}, and CHRF \citep{popovic-2015-chrf}. For GEC, they are $\rm{M^2}$ \citep{dahlmeier-ng-2012-better} and ERRANT \citep{bryant-etal-2017-automatic}. 
From the above, it is clear that these post-editing or correction tasks use reference-based evaluation metrics if manual annotation data are available. 

\section{Data Annotation}
\label{sec:data_ann}

\subsection{Source Data Selection}
We select SAMSum \citep{gliwa-etal-2019-samsum} and DialogSum \citep{chen-etal-2021-dialogsum}, the two most widely used datasets in the field of short dialogue summarization, and collect summaries generated by four systems, BART \citep{lewis-etal-2020-bart}, UniLM \citep{NEURIPS2019_c20bb2d9}, MV-BART \citep{chen-yang-2020-multi} and CODS \citep{wu-etal-2021-controllable}, on their test sets. The outputs of each system on the SAMSum test set are obtained from DialSummEval \citep{gao-wan-2022-dialsummeval}. For DialogSum, the outputs of BART and UniLM are provided by the authors of the dataset, and we retrain MV-BART and CODS on DialogSum with default settings to obtain their outputs.

We randomly sample 500 dialogues from each of the test sets of SAMSum and DialogSum, and the corresponding summaries of the above four systems, for a total of $2\times500\times4=4000$ dialogue-summary pairs, as the raw data to be annotated.

\subsection{Annotation Process}
We recruited college students as annotators. Annotators are required to be able to read and understand English daily conversations and articles fluently and have good English writing skills.

We designed the annotation interface by tagtog \footnote{\url{https://www.tagtog.com/}} to allow annotators to easily annotate multiple types of data. One dialogue and four system summaries are shown to the annotator at the same time. For each summary, the annotators will determine whether it is factually correct first. If there are factual errors in the summary, they will drag the mouse to mark the words and phrases which are factually inconsistent with the dialogue and then assign an error category by clicking the word and phrases they select. A summary may contain more than one error. Finally, if the summary contains any factual errors, they will write a corrected summary. Otherwise, the corrected summary will be the same as the original.

A detailed annotation guide was given to annotators to help them be familiar with the annotation interface and the definition of the task. Here we follow the taxonomy of factual errors proposed by \citet{pagnoni-etal-2021-understanding}. There are eight kinds of factual errors: (1) Entity Error (\textbf{EntE}); (2) Predicate Error (\textbf{PredE}); (3) Circumstance Error (\textbf{CircE}); (4) Coreference Error (\textbf{CorefE}); (5) Discourse Link Error (\textbf{LinkE}); (6) Out of Article Error (\textbf{OutE}); (7) Grammatical Error (\textbf{GramE}); (8) Others (\textbf{OthE}). Please see examples in Appendix \ref{sec:appdix_example}.

When correcting factual errors, the annotators needed to follow the three principles: (1) Correct factual errors with as few modifications as possible. (2) Making substitutions for words and phrases is preferred. When substitution is difficult, deletion can be performed. (3) The corrected summary should be grammatically correct, coherent, and non-redundant as possible.


\begin{figure*}[t]
\centering
\includegraphics[width=\textwidth,height=0.2\textwidth]{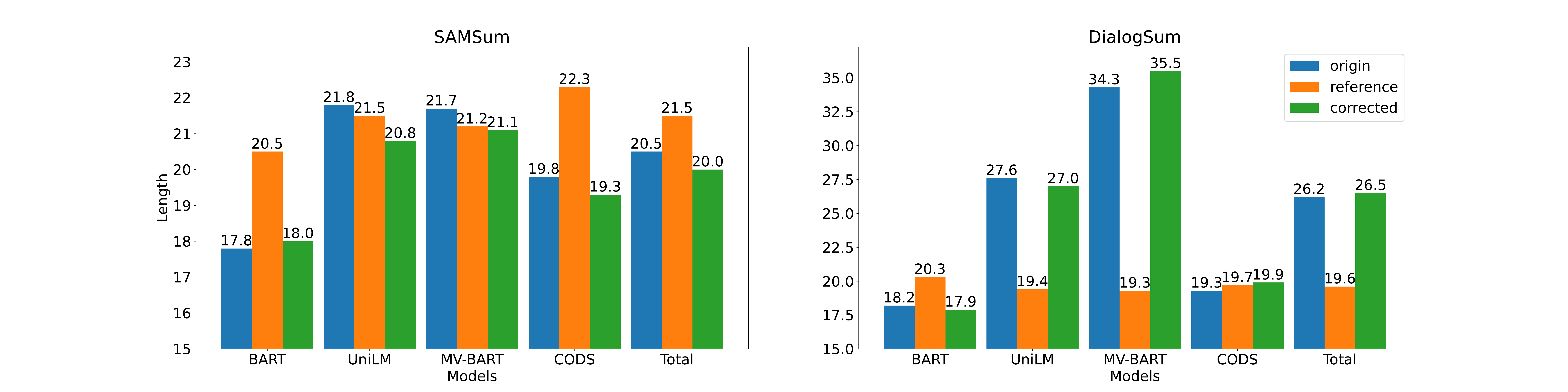}
\caption{The average length of original summaries, reference summaries, and corrected summaries. Only items with factual errors in the original summary are counted.}
\label{fig:datasets_length}
\end{figure*}

We divided the original data into 10 batches, each containing 100 dialogues ($100\times4=400$ items). In order to ensure annotation quality, those who wished to participate in the annotation were required to complete the annotation of all the summaries corresponding to the 10 dialogues ($10\times4=40$ items) first. After completing this small part, we evaluated the annotation results, pointed out any inappropriate annotations, and told them our suggestions. After confirming that the annotation task was correctly understood, the participants were allowed to continue annotation. In subsequent annotation, we sampled the results to check. Throughout the process, we kept in touch with the annotators via email and instant messaging software.

\subsection{Data Analysis}


It is necessary to illustrate the difference between the manually annotated corrected summaries and the reference summaries in the dialogue summarization dataset. We focus on their relationship to the summaries to be corrected. Since the summaries that do not contain factual errors do not need to be corrected, i.e., the corrected summaries are the same as the original summaries, we only count data for samples where the original summaries contain factual errors. For these samples, it can be seen from Figure \ref{fig:datasets_length} that the corrected summaries are closer in length to the original summaries compared to the reference summaries. This is more obvious on DialogSum. As shown in Table \ref{tab:datasets_bleu}, the corrected summaries are closer to the original summaries in terms of n-gram overlap compared to the reference summaries. This result is in line with our annotation principles.

\begin{table}
\centering
\resizebox{\linewidth}{!}{
\begin{tabular}{cccccc}
\hline & \textbf{BART} & \textbf{UniLM} & \textbf{MV-BART} & \textbf{CODS} & \textbf{Total} \\
\hline
\textbf{SAMSum} & 0.43 / 0.85 & 0.39 / 0.82 & 0.45 / 0.85 & 0.46 / 0.84 & 0.43 / 0.84 \\
\textbf{DialogSum} & 0.62 / 0.73 & 0.54 / 0.68 & 0.56 / 0.76 & 0.61 / 0.72 & 0.58 / 0.72
 \\
\hline
\end{tabular}
}
\caption{BLEU score comparison (origin vs. reference / origin vs. corrected). Only items with factual errors in the original summary are counted. }
\label{tab:datasets_bleu}
\end{table}

\begin{table}
\centering
\resizebox{\linewidth}{!}{
\begin{tabular}{cccccc}
\hline & \textbf{BART} & \textbf{UniLM} & \textbf{MV-BART} & \textbf{CODS} & \textbf{Total} \\
\hline
\textbf{SAMSum} & 26.00 & 51.20 & 37.00 & 44.40 & 39.65 \\
\textbf{DialogSum} & 31.20 & 44.80 & 58.00 & 40.60 & 43.65 \\
\hline
\end{tabular}
}
\caption{Percentage of summaries with factual errors.}
\label{tab:dataset_fact_error}
\vspace{-5mm}
\end{table}

\begin{figure}
\centering
\includegraphics[width=0.5\textwidth,height=0.2\textwidth]{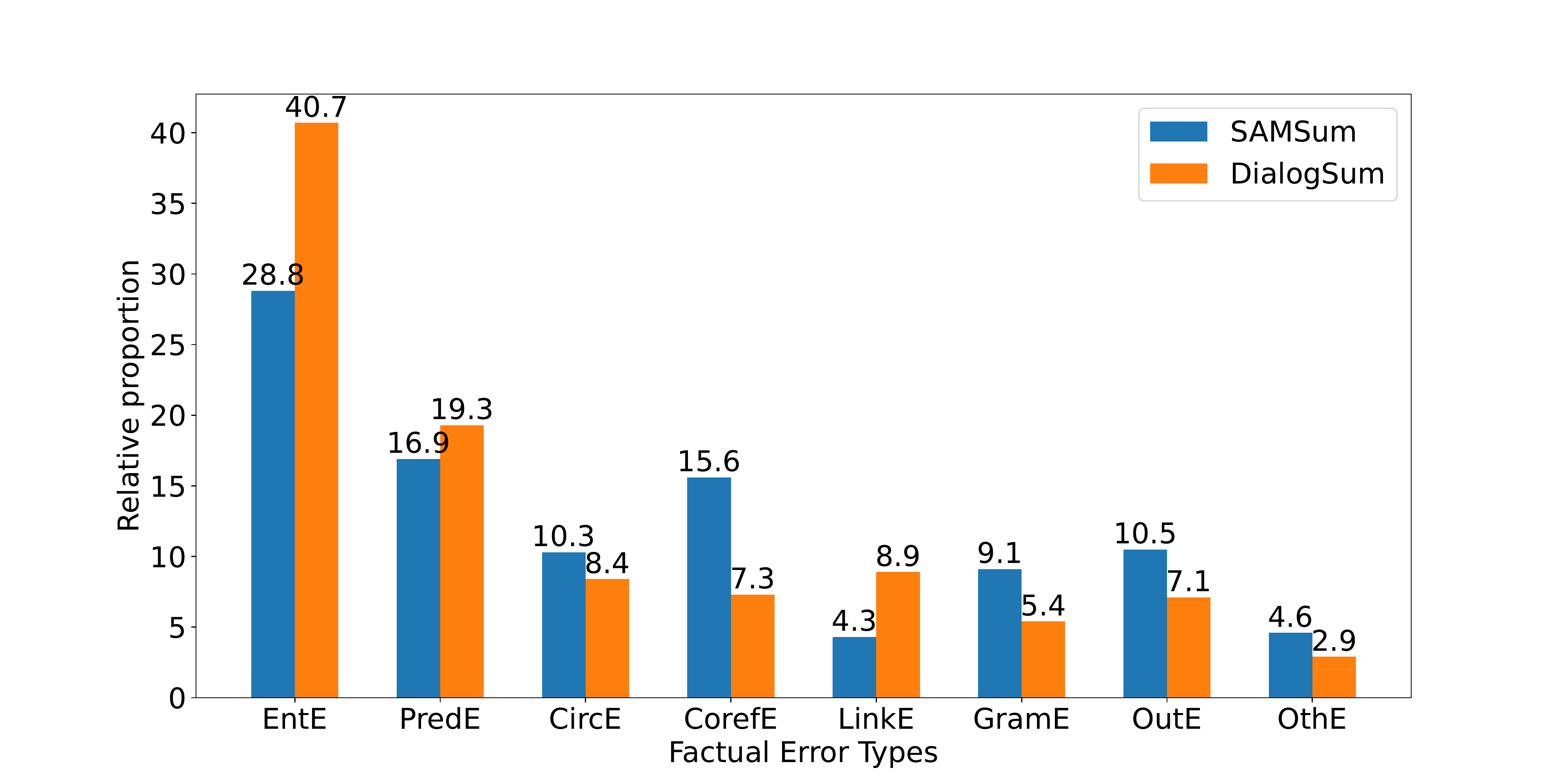}
\caption{The relative percentage of factual error types. A factually incorrect span is counted as one error.}
\label{fig:error_proportion}
\end{figure}
 
For the percentage of factual inconsistencies and error categories, as shown in Table \ref{tab:dataset_fact_error}, around 40\% of generated summaries contain factual errors. This ratio is similar to the annotation results of \citet{wang2022analyzing}. Figure \ref{fig:error_proportion} shows that, \textbf{EntE} and \textbf{PredE} are the two most dominant types of errors. It is important to note that the percentage of \textbf{GramE} (difficult to understand due to grammatical errors) is less. This is in line with the findings of \citet{gao-wan-2022-dialsummeval}: the current dialogue summarization systems based on pre-trained models generate summaries that are already good in terms of fluency.

\begin{table*}[]
\resizebox{\linewidth}{!}{
\begin{tabular}{crrrrr|crrrrr}
\hline
\multicolumn{6}{c|}{\textbf{SAMSum (N=793)}} & \multicolumn{6}{c}{\textbf{DialogSum (N=873)}} \\ \hline
\multicolumn{1}{l|}{} & \multicolumn{1}{c}{\textbf{origin}} & \multicolumn{1}{c|}{\textbf{correct}} & \multicolumn{1}{c}{\textbf{\textless{}}} & \multicolumn{1}{c}{\textbf{=}} & \multicolumn{1}{c|}{\textbf{\textgreater{}}} & \multicolumn{1}{c|}{\textbf{}} & \multicolumn{1}{c}{\textbf{origin}} & \multicolumn{1}{c|}{\textbf{correct}} & \multicolumn{1}{c}{\textbf{\textless{}}} & \multicolumn{1}{c}{\textbf{=}} & \multicolumn{1}{c}{\textbf{\textgreater{}}} \\ \hline
\multicolumn{1}{c|}{\textbf{FactCC}} & 0.136 & \multicolumn{1}{r|}{0.139} & 0.04 & 0.93 & 0.03 & \multicolumn{1}{c|}{\textbf{FactCC}} & 0.286 & \multicolumn{1}{r|}{0.276} & 0.04 & 0.91 & 0.05 \\ \hline
\multicolumn{1}{c|}{\textbf{DAE}} & 0.076 & \multicolumn{1}{r|}{0.077} & 0.02 & 0.97 & 0.02 & \multicolumn{1}{c|}{\textbf{DAE}} & 0.199 & \multicolumn{1}{r|}{0.207} & 0.04 & 0.93 & 0.03 \\ \hline
\multicolumn{1}{c|}{\textbf{QuestEval}} & 0.392 & \multicolumn{1}{r|}{0.380} & 0.30 & 0.23 & 0.47 & \multicolumn{1}{c|}{\textbf{QuestEval}} & 0.486 & \multicolumn{1}{r|}{0.486} & 0.34 & 0.31 & 0.34 \\ \hline
\multicolumn{1}{c|}{\textbf{BARTScore}} & -3.084 & \multicolumn{1}{r|}{-3.123} & 0.25 & 0.53 & 0.22 & \multicolumn{1}{c|}{\textbf{BARTScore}} & -2.826 & \multicolumn{1}{r|}{-2.810} & 0.21 & 0.57 & 0.22 \\ \hline
\end{tabular}
}
\caption{Test results of factuality metrics. The higher the output scores of the metrics, the better the factuality. Only summaries with factual errors are counted. \textbf{origin} and \textbf{correct} refer to the average of the output scores of the original summaries and the corrected summaries. \textbf{<}, \textbf{=}, and \textbf{>} refer to the proportion of scores of the original summary that are less than, equal to, and greater than that of the corrected summary, when compared as a pair. }
\label{tab:factual_test}
\end{table*}

\section{Test for Factuality Metrics}
\label{sec:test_facutal}
We perform a simple test of the reliability of factuality metrics using the above dataset. In general, the factuality metric $F$ takes the source document $S$ and the summary $H$ as inputs and outputs a score $F(S,H)$. A reliable factual indicator needs to satisfy the condition that for summaries with factual errors, the factual score of the corrected summary $C$ is greater than that of the original summary $O$,
i.e., $F(S, C) > F(S, O)$.

We select four commonly used factuality metrics: FactCC \citep{kryscinski-etal-2020-evaluating}, DAE (\citealp{goyal-durrett-2020-evaluating}; \citealp{goyal-durrett-2021-annotating}), QuestEval \citep{scialom-etal-2021-questeval}, and BARTScore \citep{yuan2021bartscore}. Table \ref{tab:factual_test} illustrates that it is unreliable to evaluate the factuality of the original and corrected summaries using these metrics. The factuality scores of the corrected summaries are not significantly better than those of the original summaries under these metrics, either in mean or pairwise comparisons.

\begin{figure*}
\centering
\includegraphics[width=1.0\textwidth,height=0.3\textwidth]{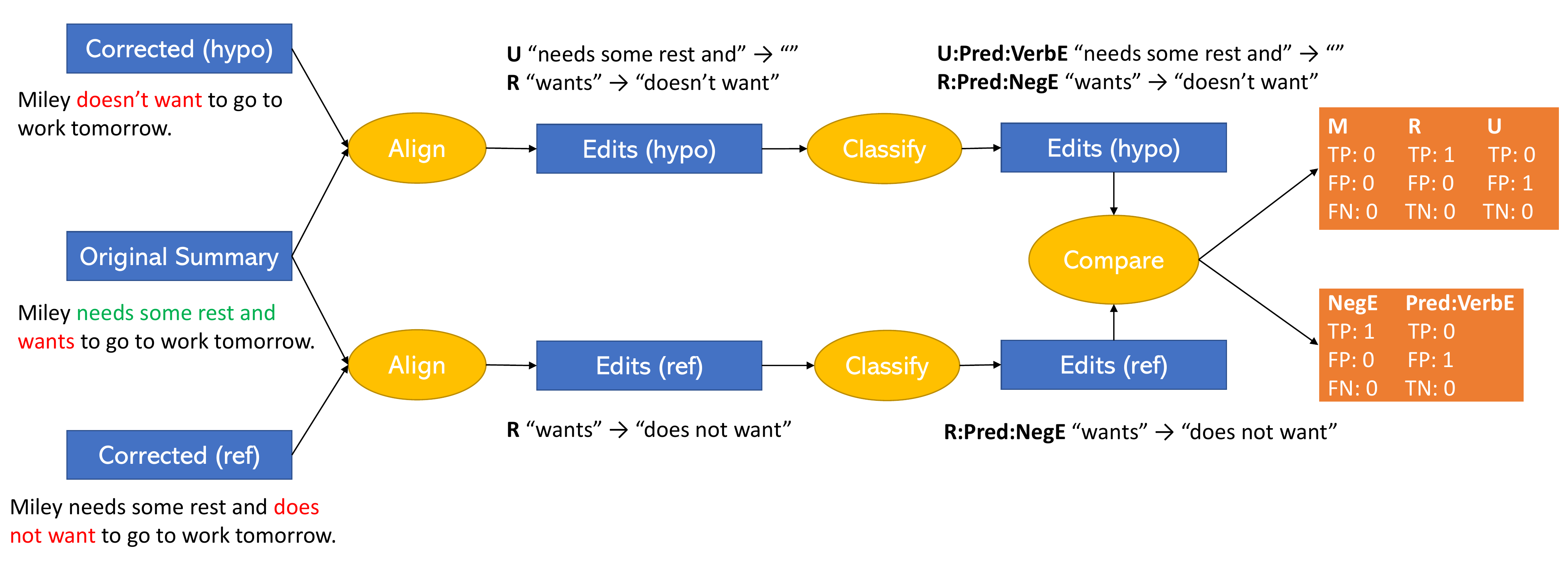}
\caption{Diagram of our evaluation framework, FERRANTI. \textcolor{red}{Red} parts indicate replacement (R). \textcolor{green}{Green} parts indicate deletion (U). Addition edits do not appear in this example (M).}
\label{fig:framework}
\vspace{-5mm}
\end{figure*}

\section{Reference-based Evaluation Framework}
\label{sec:eval_frame}
We find that 
it is difficult for manual annotation to determine the boundaries of erroneous spans accurately sometimes, which hinders the fine-grained evaluation of FEC models by error categories.
Considering these error categories have clear linguistic characteristics, it is more feasible to use a rule-based approach to automatically align and determine error categories when reference correction is already available.

We propose FERRANTI, a \textbf{F}actual \textbf{ERR}or \textbf{AN}notation \textbf{T}oolk\textbf{I}t designed for FEC. Noting the great practice of ERRANT \citep{bryant-etal-2017-automatic}, our implementation builds on it. As shown in Figure \ref{fig:framework}, it mainly consists of three steps: alignment, classification, and comparison.

\subsection{Taxonomy of Factual Errors}
To automatically classify factual errors for FEC, we propose a new taxonomy of factual errors. Compared to existing classifications of factual errors, such as \citet{pagnoni-etal-2021-understanding}, \citet{tang-etal-2022-confit} and \citet{wang2022analyzing}, our taxonomy differs in three main ways: (1) we point out that there are two classifications of factual errors of different perspectives, content-based and form-based; (2) we hierarchize the content-based classification of factual errors; (3) our error classification is implemented by explicit linguistic rules rather than manual annotation.

The content-based categories are shown in Table \ref{tab:content_based}. In this classification, the category to which an edit belongs needs to draw on the POS of the words in the sentence as well as on the dependencies. Compared to the classification we used in the annotation, we subdivide \textbf{EntE} and \textbf{PredE}, add \textbf{NumE}, and do not use \textbf{OutE} and \textbf{GramE} that have unclear POS and dependency features. By this, we cover special categories such as negation errors (\textbf{NegE}) that received attention in summarization factuality without losing generality. 

The form-based categories are shown in Table \ref{tab:form_based}. They are called form-based because, in factual error correction, it is basically only necessary to align the original summary and the corrected summary by whether the words are the same to determine whether an edit is an addition, deletion, or modification. \citet{devaraj-etal-2022-evaluating} adopt a similar way when analyzing the factuality of text simplification.

It is necessary to point out that the form-based and content-based classifications are not mutually exclusive. They can be combined, such as \textbf{R:Pred:Neg} in Figure \ref{fig:framework}.

\begin{table*}
\centering
\resizebox{\linewidth}{!}{
\begin{tabular}{|c|c|c|c|}
\hline
\textbf{Code} & \textbf{Meaning} & \textbf{Description} & \textbf{Examples} \\
\hline
\textbf{M} & Missing & Missing information that needs to be added. & \textit{with Ms.} $\rightarrow$ \textit{with Ms. Blair}
 \\
\hline
\textbf{R} & Replacement & Wrong information that needs to be modified. & 
\textit{reminds} $\rightarrow$ \textit{teaches} \\
\hline
\textbf{U} & Unnecessary & Redundant information that needs to be deleted. & \textit{Derek and Phil} $\rightarrow$ \textit{Derek} \\
\hline
\end{tabular}
}
\caption{Form-based categories of factual errors.}
\label{tab:form_based}
\end{table*}

\begin{table*}
\centering
\resizebox{\linewidth}{!}{
\begin{tabular}{|l|p{10cm}|l|}
\hline
\textbf{Code} & \textbf{Description} & \textbf{Example} \\
\hline
\textbf{Ent:ObjE} & Object errors in entity errors, mainly nouns. & \textit{Laura} $\rightarrow$ \textit{Paul} \\
\hline
\textbf{Ent:AttrE} & Attribute errors in entity errors, mainly adjectives. & \textit{proud} $\rightarrow$ \textit{happy} \\
\hline
\textbf{Pred:ModE} & Modality errors in predicate errors, mainly modal verbs that express possibilities. & \textit{is} $\rightarrow$ \textit{may be} \\
\hline
\textbf{Pred:TensE} & Tense errors in predicate errors. & \textit{is} $\rightarrow$ \textit{was} \\
\hline
\textbf{Pred:NegE} & Negation errors in predicate errors. & \textit{will} $\rightarrow$ \textit{won't} \\
\hline
\textbf{Pred:VerbE} & General predicate errors that do not fall into the above categories. & \textit{lent} $\rightarrow$ \textit{gave} \\
\hline
\textbf{CircE} & Circumstance errors, mainly adverbs, prepositional phrases, etc. & \textit{after} $\rightarrow$ \textit{during} \\
\hline
\textbf{CorefE} & Coreference errors, mainly pronouns. & \textit{her} $\rightarrow$ \textit{Ann} \\
\hline
\textbf{LinkE} & Link errors, conjunctions & \textit{but} $\rightarrow$ \textit{because} \\
\hline
\textbf{NumE} & Errors in numbers & \textit{15} $\rightarrow$ \textit{30} \\
\hline
\textbf{OthE} & Other errors that are not all of the above types of errors. & \textit{, so she} $\rightarrow$ \textit{. She}  \\
\hline
\end{tabular}
}
\caption{Content-based categories of factual errors. The examples in the table are all replacements, but deletions and additions are also possible.}
\label{tab:content_based}
\vspace{-5mm}
\end{table*}

\subsection{Alignment}
In this step, the corrected summaries are aligned with the original ones and the edits are extracted automatically. We follow ERRANT by using an alignment algorithm that considers linguistic features as a cost function \citep{felice-etal-2016-automatic}. However, unlike ERRANT, we merge all adjacent edits considering that a small number of factually corrected edits are longer. Before alignment, the summary is pre-processed with Spacy\footnote{version 2.3.0, \url{https://spacy.io/}} for tokenization, POS tagging, etc. Form-based error categories are automatically assigned to each edit after alignment.

\subsection{Classification}
After edits are extracted, they are assigned content-based categories based on the linguistic features of the original span and the corrected span (mainly POS and lemma). The detailed rules are not listed here. 

\subsection{Comparison}
In this step, hypothesis edits and reference edits are compared and scores are computed in different categories for form-based and content-based categories. 
Edits that appear in both hypothesis and reference are true positive (\textbf{TP}). For TP, we use the category of edits in reference as the final category. Edits that appear only in the hypothesis or reference are false positive (\textbf{FP}) or false negative (\textbf{FN}). Further, we can obtain precision, recall, and F-values. We report $\rm{F_{0.5}}$ out of a penalty for over-correction.

\section{Experiments}

\subsection{FEC approaches}
We select a few representative FEC approaches. Among them, we are most interested in such methods: generating corrected summaries directly based on data augmentation because of their flexibility. 


\textbf{Rule-based transformation} \citet{cao-etal-2020-factual} use a set of rules that swap the entities, numbers, dates, and pronouns of the reference summaries to construct the summaries to be corrected for training. We call this approach \textbf{rule}.

\textbf{Infilling-based transformation} \citet{Balachandran2022CorrectingDF} mask and predict the subjects, relations, and objects of sentences in the source documents to train an infilling model. The reference summaries are then masked in the same way, and the trained infilling model is used to fill the masked reference summaries to construct the summaries to be corrected. 
For the infilling model, we experiment with two different setups: (1) using the trained infilling model from the original study, denoted as \textbf{infill}. (2) retraining the infilling model 
, denoted as \textbf{infill-r}. Please see Appendix \ref{sec:appendix_retrain} for the details of retraining. 

In addition to the method of generating a corrected summary directly, we also select other approaches, which aim at correcting extrinsic hallucinations:

\textbf{CCGS} \citet{chen-etal-2021-improving} replace named entities and numbers in reference summary with the compatible semantic type of content from the source document to generate candidates to train a factual classifier based on BART. At the time of inference, the candidates for the summary to be corrected are generated in a similar way, the trained classifier is used to re-rank the candidates, and the best one is selected as the corrected summary.

\begin{table*}[t]
\resizebox{\linewidth}{!}{

\begin{tabular}{|cccccc|}
\hline
\multicolumn{6}{|c|}{\textbf{SAMSum}}                              \\ \hline
\multicolumn{6}{|c|}{BART as the pre-trained model}                                                                                     \\ \hline
\multicolumn{1}{|c|}{\textbf{Type}} & \textbf{Pseudo} & \textbf{Real} & \textbf{Pseudo+Real} & \textbf{RefS} & \textbf{Pseudo+RefS} \\ \hline
\multicolumn{1}{|c|}{\textbf{M}} & 0.00 & 0.00 & 0.00 & \textbf{2.08} & 2.02                 \\
\multicolumn{1}{|c|}{\textbf{R}} & 4.26 & 7.58 & \textbf{15.00} & 2.34 & 1.44                 \\
\multicolumn{1}{|c|}{\textbf{U}} & 7.04 & 6.07 & \textbf{13.66} & 3.89 & 4.66                 \\
\multicolumn{1}{|c|}{\textbf{Total}} & 4.15 & 5.63 & \textbf{13.01} & 2.54 & 2.33                 \\ \hline
\multicolumn{6}{|c|}{PEGASUS as pre-trained   models}                                                                                \\ \hline
\multicolumn{1}{|c|}{\textbf{Type}} & \textbf{Pseudo} & \textbf{Real} & \textbf{Pseudo+Real} & \textbf{RefS} & \textbf{Pseudo+RefS} \\ \hline
\multicolumn{1}{|c|}{\textbf{M}} & 0.00 & 0.00 & 0.00 & 1.41 & \textbf{1.59}        \\
\multicolumn{1}{|c|}{\textbf{R}} & 12.15 & 1.58 & \textbf{13.72} & 2.58 & 4.68                 \\
\multicolumn{1}{|c|}{\textbf{U}} & \textbf{7.46} & 4.05 & 7.04 & 1.17 & 4.25                 \\
\multicolumn{1}{|c|}{\textbf{Total}} & 9.48 & 2.15 & \textbf{10.82} & 1.99 & 4.18                 \\ \hline
\multicolumn{6}{|c|}{T5 as the pre-trained model}                                                                                       \\ \hline
\multicolumn{1}{|c|}{\textbf{Type}} & \textbf{Pseudo} & \textbf{Real} & \textbf{Pseudo+Real} & \textbf{RefS} & \textbf{Pseudo+RefS} \\ \hline
\multicolumn{1}{|c|}{\textbf{M}} & 0.00 & 0.00 & 0.00 & 0.00 & \textbf{0.88}        \\
\multicolumn{1}{|c|}{\textbf{R}} & 10.54 & 0.00 & \textbf{16.18} & 3.52 & 4.66                 \\
\multicolumn{1}{|c|}{\textbf{U}} & 7.94 & 18.99 & \textbf{24.10} & 6.26 & 7.99                 \\
\multicolumn{1}{|c|}{\textbf{Total}} & 8.89 & 4.72 & \textbf{15.69} & 3.74 & 4.57                 \\ \hline
\end{tabular}

\begin{tabular}{|cccccc|}
\hline
\multicolumn{6}{|c|}{\textbf{DialogSum}}                              \\ \hline
\multicolumn{6}{|c|}{BART as the pre-trained model}                              \\ \hline
\multicolumn{1}{|c|}{\textbf{Type}} & \textbf{Pseudo} & \textbf{Real} & \textbf{Pseudo+Real} & \textbf{RefS} & \textbf{Pseudo+RefS} \\ \hline
\multicolumn{1}{|c|}{\textbf{M}} & \textbf{5.49} & 0.00 & 0.00 & 0.75 & 2.32                 \\
\multicolumn{1}{|c|}{\textbf{R}} & \textbf{3.48} & 1.72 & 1.74 & 1.58 & 1.34                 \\
\multicolumn{1}{|c|}{\textbf{U}} & \textbf{12.05} & 4.32 & 4.57 & 3.02 & 2.43                 \\
\multicolumn{1}{|c|}{\textbf{Total}} & \textbf{4.24} & 2.43 & 2.31 & 1.76 & 1.66                 \\ \hline
\multicolumn{6}{|c|}{PEGASUS as pre-trained   models}                          \\ \hline
\multicolumn{1}{|c|}{\textbf{Type}} & \textbf{Pseudo} & \textbf{Real} & \textbf{Pseudo+Real} & \textbf{RefS} & \textbf{Pseudo+RefS} \\ \hline
\multicolumn{1}{|c|}{\textbf{M}} & \textbf{14.93} & 0.00 & 0.00 & 0.00 & 0.78                 \\
\multicolumn{1}{|c|}{\textbf{R}} & \textbf{9.32} & 5.75 & 4.44 & 2.10 & 2.19                 \\
\multicolumn{1}{|c|}{\textbf{U}} & \textbf{13.33} & 3.70 & 0.00 & 2.84 & 1.41                 \\
\multicolumn{1}{|c|}{\textbf{Total}} & \textbf{10.25} & 4.58 & 3.50 & 1.98 & 1.87                 \\ \hline
\multicolumn{6}{|c|}{T5 as the pre-trained model}                                    \\ \hline
\multicolumn{1}{|c|}{\textbf{Type}} & \textbf{Pseudo} & \textbf{Real} & \textbf{Pseudo+Real} & \textbf{RefS} & \textbf{Pseudo+RefS} \\ \hline
\multicolumn{1}{|c|}{\textbf{M}} & \textbf{13.33} & 0.00 & 0.00 & 1.45 & 3.26                 \\
\multicolumn{1}{|c|}{\textbf{R}} & 7.33 & 1.35 & \textbf{8.29} & 2.46 & 4.26                 \\
\multicolumn{1}{|c|}{\textbf{U}} & 7.46 & 16.95 & \textbf{18.18} & 3.36 & 4.18                 \\
\multicolumn{1}{|c|}{\textbf{Total}} & 7.89 & 2.98 & \textbf{8.33} & 2.50 & 4.12                 \\ \hline
\end{tabular}
}

\caption{Performance (FERRANTI: form-based categories defined in Table \ref{tab:form_based}) of different training modes on SAMSum and DialogSum. The values are all $\rm{F_{0.5}}$ scores. The best results under the same pre-trained model are bolded. The data augmentation approach is set to \textbf{rule}. The pseudo data for the two datasets are constructed separately. Please see Table \ref{tab:train_mode_samsum_form}, Table \ref{tab:train_mode_samsum_tp_form}, Table \ref{tab:train_mode_dialogsum_form}, and Table \ref{tab:train_mode_dialogsum_tp_form} in Appendix \ref{sec:appendix_fig_tab} for precision, recall, or TP, etc. }
\label{tab:train_mode_samsum_dialogsum_short_form}
\end{table*}

\textbf{FactPegasus}  \citet{wan-bansal-2022-factpegasus} propose a component for correcting factual errors without training data: based on manually written rules and the Spacy library, and it removes or replaces entities and related content in the summary that do not appear in the source document.


\subsection{Training Modes}
For different data augmentation approaches (\textbf{rule}, \textbf{infill}, and \textbf{infill-r}), we conduct experiments with different training modes to explore some factors of interest. To compare the role played by pseudo data (generated by data augmentation)  and real data (manually annotated) in training, we designed the following training modes: (1) Training with pseudo data only (\textbf{Pseudo}). (2) Training with real data only (\textbf{Real}). (3) Training with pseudo data first, then with real data (\textbf{Pseudo + Real}). In order to compare the difference between the reference correction and the reference summary of the summarization dataset, we also design the following training modes: (4) Replace the reference correction in the real data with the reference summary for training (\textbf{RefS}). (5) Training with pseudo data first, then proceed to (4) (\textbf{Pseduo + RefS}).

\subsection{Datasets and Settings}
We split our annotated dataset (which we call the real data) into a training set, a validation set, and a test set. Specifically, for the 500 dialogues of SAMSum, we split them according as 300/100/100. Each dialogue has the corresponding four model-generated original summaries and corrected summaries. The total size is 1200/400/400. For the 500 dialogue of DialogSum, the split is the same as SAMSum. We train and test models separately on the two parts (datasets).
Please see Appendix \ref{sec:appendix_model_set} for model settings and training details.

\subsection{Evaluation}
We use the evaluation framework presented in Section \ref{sec:eval_frame}, FERRANTI to automatically evaluate FEC approaches on the test set. For comparison, we also adopt factuality metrics mentioned in Section \ref{sec:test_facutal}. 

\section{Results and Analysis}

\begin{table*}[]
\resizebox{\linewidth}{!}{

\begin{tabular}{|c|rrrrrr|rrrrrr|rrrrrr|}
\hline
\multicolumn{1}{|l|}{\multirow{3}{*}{}} & \multicolumn{6}{c|}{BART as the pre-trained model} & \multicolumn{6}{c|}{PEGASUS as the pre-trained model} & \multicolumn{6}{c|}{T5 as the pre-trained model} \\ \cline{2-19} 
\multicolumn{1}{|l|}{} & \multicolumn{3}{c}{\textbf{Pseudo}} & \multicolumn{3}{c|}{\textbf{Pseudo+Real}} & \multicolumn{3}{c}{\textbf{Pseudo}} & \multicolumn{3}{c|}{\textbf{Pseudo+Real}} & \multicolumn{3}{c}{\textbf{Pseudo}} & \multicolumn{3}{c|}{\textbf{Pseudo+Real}} \\ \cline{2-19} 
\multicolumn{1}{|l|}{} & \multicolumn{1}{c}{\textbf{rule}} & \multicolumn{1}{c}{\textbf{infill}} & \multicolumn{1}{c}{\textbf{infill-r}} & \multicolumn{1}{c}{\textbf{rule}} & \multicolumn{1}{c}{\textbf{infill}} & \multicolumn{1}{c|}{\textbf{infill-r}} & \multicolumn{1}{c}{\textbf{rule}} & \multicolumn{1}{c}{\textbf{infill}} & \multicolumn{1}{c}{\textbf{infill-r}} & \multicolumn{1}{c}{\textbf{rule}} & \multicolumn{1}{c}{\textbf{infill}} & \multicolumn{1}{c|}{\textbf{infill-r}} & \multicolumn{1}{c}{\textbf{rule}} & \multicolumn{1}{c}{\textbf{infill}} & \multicolumn{1}{c}{\textbf{infill-r}} & \multicolumn{1}{c}{\textbf{rule}} & \multicolumn{1}{c}{\textbf{infill}} & \multicolumn{1}{c|}{\textbf{infill-r}} \\ \hline
\textbf{Ent:ObjE} & 2.02 & \textbf{11.36} & 4.72 & 10.59 & 9.09 & 9.62 & 14.71 & 5.21 & 8.93 & \textbf{16.83} & 11.90 & 9.62 & 14.34 & 4.03 & 3.47 & \textbf{16.20} & 3.47 & 7.46 \\ \hline
\textbf{Ent:AttrE} & 0.00 & 0.00 & 0.00 & 0.00 & 0.00 & 0.00 & 0.00 & 0.00 & 0.00 & 0.00 & 0.00 & 0.00 & 0.00 & 0.00 & 0.00 & 0.00 & 0.00 & 0.00 \\ \hline
\textbf{Pred:ModE} & 0.00 & 0.00 & 0.00 & 0.00 & 0.00 & 0.00 & 0.00 & 0.00 & 0.00 & 0.00 & 0.00 & 0.00 & 0.00 & 0.00 & 0.00 & 0.00 & 0.00 & 0.00 \\ \hline
\textbf{Pred:TensE} & - & - & 0.00 & 0.00 & - & - & - & 0.00 & 0.00 & - & 0.00 & - & - & - & - & - & - & - \\ \hline
\textbf{Pred:NegE} & 0.00 & 0.00 & 0.00 & 0.00 & 0.00 & 0.00 & 0.00 & 0.00 & 0.00 & 0.00 & 0.00 & 0.00 & 0.00 & 0.00 & 0.00 & 0.00 & 0.00 & 0.00 \\ \hline
\textbf{Pred:VerbE} & 4.76 & 2.65 & 2.70 & \textbf{12.12} & 7.30 & 11.30 & 0.00 & 3.76 & 3.18 & 0.00 & \textbf{9.52} & 4.59 & 0.00 & 0.00 & 0.00 & 13.76 & 12.00 & \textbf{14.29} \\ \hline
\textbf{CircE} & 0.00 & 0.00 & 0.00 & 0.00 & 0.00 & 0.00 & 0.00 & 0.00 & 0.00 & 0.00 & 0.00 & 0.00 & 0.00 & 0.00 & 0.00 & 0.00 & 0.00 & 0.00 \\ \hline
\textbf{CorefE} & 7.04 & 0.00 & 0.00 & \textbf{25.32} & 10.99 & 7.04 & 0.00 & 0.00 & 0.00 & 5.75 & 6.67 & \textbf{17.24} & 7.04 & 0.00 & 0.00 & \textbf{24.27} & 6.02 & 7.46 \\ \hline
\textbf{LinkE} & 0.00 & 0.00 & 0.00 & 0.00 & 0.00 & 0.00 & 0.00 & 0.00 & 0.00 & 0.00 & 0.00 & 0.00 & 0.00 & 0.00 & 0.00 & 0.00 & 0.00 & 0.00 \\ \hline
\textbf{NumE} & 31.25 & 0.00 & 0.00 & \textbf{41.67} & 0.00 & 0.00 & 41.67 & 0.00 & 0.00 & \textbf{50.00} & 0.00 & 0.00 & 0.00 & 0.00 & 0.00 & 0.00 & 0.00 & 0.00 \\ \hline
\textbf{OthE} & - & 0.00 & - & - & - & - & 0.00 & 0.00 & 0.00 & 0.00 & 0.00 & 0.00 & 0.00 & 0.00 & 0.00 & 0.00 & 0.00 & 0.00 \\ \hline
\textbf{Total} & 4.15 & 5.23 & 2.92 & \textbf{13.01} & 8.10 & 8.96 & 9.48 & 3.33 & 4.44 & \textbf{10.82} & 8.54 & 8.62 & 8.89 & 1.57 & 1.27 & \textbf{15.69} & 6.28 & 6.98 \\ \hline
\end{tabular}
}
\caption{Performance (FERRANTI: content-based categories defined in Table \ref{tab:content_based} of different data augmentation approaches on SAMSum. The results on DialogSum are shown in Table \ref{tab:approach_dialogsum_short_content} in Appendix \ref{sec:appendix_fig_tab}. The values are all $\rm{F_{0.5}}$ scores. The best results under the same pre-trained model are bolded. The pseudo-data corpus is SAMSum. Please see Table \ref{tab:approach_samsum_content} and Table \ref{tab:approach_samsum_tp_content} in Appendix \ref{sec:appendix_fig_tab} for precision, recall, or TP, etc.}
\label{tab:approach_samsum_short_content}
\end{table*}

\begin{table}[]
\resizebox{\linewidth}{!}{
\begin{tabular}{|crrr|crrr|}
\hline
\multicolumn{4}{|c|}{\textbf{SAMSum}} & \multicolumn{4}{c|}{\textbf{DialogSum}} \\ \hline
\multicolumn{1}{|c|}{Pre-trained Models} & \multicolumn{1}{c}{BART} & \multicolumn{1}{c}{BERT} & \multicolumn{1}{c|}{RoBERTa} & \multicolumn{1}{c|}{Pre-trained Models} & \multicolumn{1}{c}{BART} & \multicolumn{1}{c}{BERT} & \multicolumn{1}{c|}{RoBERTa} \\ \hline
\multicolumn{1}{|c|}{\textbf{M}} & 0.00 & 0.00 & 0.00 & \multicolumn{1}{c|}{\textbf{M}} & 0.00 & 0.00 & 0.00 \\ 
\multicolumn{1}{|c|}{\textbf{R}} & 1.98 & 0.00 & 1.52 & \multicolumn{1}{c|}{\textbf{R}} & 1.06 & 2.22 & 2.42 \\ 
\multicolumn{1}{|c|}{\textbf{U}} & 0.00 & 0.00 & 0.00 & \multicolumn{1}{c|}{\textbf{U}} & 0.00 & 0.00 & 0.00 \\ 
\multicolumn{1}{|c|}{\textbf{Total}} & 1.41 & 0.00 & 1.14 & \multicolumn{1}{c|}{\textbf{Total}} & 0.87 & 1.80 & 1.91 \\ \hline
\end{tabular}
}
\caption{Performance (FERRANTI: form-based categories) of CCGS on SAMSum and DialogSum.} 
\label{tab:ccgs_samsum_dialogsum_short_form}
\end{table}

\begin{table}[]
\resizebox{\linewidth}{!}{
\begin{tabular}{|cccc|cccc|}
\hline
\multicolumn{4}{|c|}{\textbf{SAMSum}} & \multicolumn{4}{c|}{\textbf{DialogSum}} \\ \hline
\multicolumn{1}{|c|}{Spacy Models} & sm & md & lg & \multicolumn{1}{c|}{Spacy Models} & sm & md & lg \\ \hline
\multicolumn{1}{|c|}{\textbf{M}} & 0.00 & 0.00 & 0.00 & \multicolumn{1}{c|}{\textbf{M}} & 0.00 & 0.00 & 0.00 \\
\multicolumn{1}{|c|}{\textbf{R}} & 0.00 & 0.00 & 0.00 & \multicolumn{1}{c|}{\textbf{R}} & 0.00 & 0.00 & 0.00 \\
\multicolumn{1}{|c|}{\textbf{U}} & 1.80 & 1.62 & 3.99 & \multicolumn{1}{c|}{\textbf{U}} & 0.58 & 1.32 & 0.36 \\
\multicolumn{1}{|c|}{\textbf{Total}} & 1.42 & 1.21 & 2.98 & \multicolumn{1}{c|}{\textbf{Total}} & 0.37 & 0.91 & 0.26 \\ \hline
\end{tabular}
}
\caption{Performance (FERRANTI: form-based categories) of FactPegasus on SAMSum and DialogSum.} 
\label{tab:factpegasus_samsum_dialogsum_short_form}
\end{table}

\subsection{Performance across Training Modes}
Here we show the results on the category of form-based errors. Content-based results are shown in Table \ref{tab:train_mode_samsum_content} and Table \ref{tab:train_mode_dialogsum_content} in Appendix \ref{sec:appendix_fig_tab}.

\textbf{Reference summary vs. Reference correction} Table \ref{tab:train_mode_samsum_dialogsum_short_form} illustrates that in most cases where FERRANTI is used as the evaluation framework, training FEC models using the reference summary as the final correction target (\textbf{RefS}, \textbf{Pseudo+RefS}) does not yield good results. Tables \ref{tab:train_mode_samsum_tp_form} and \ref{tab:train_mode_dialogsum_tp_form} in Appendix \ref{sec:appendix_fig_tab} illustrate that both modes present many FPs on various error types, i.e., false edits. This is to be expected since we have shown in Section \ref{sec:data_ann} that there is a large difference between the reference correction and the reference summary. Interestingly, if evaluated using factuality metrics, we find that training with the reference summary gives the best results in most cases (the results are shown in Table \ref{tab:train_mode_samsum_dialogsum_fact} in Appendix \ref{sec:appendix_fig_tab}). This suggests that it is essential to introduce reference correction in FEC evaluation. Otherwise, FEC for summarization may lose its meaning, since the seemingly best results can be obtained by using reference summaries unrelated to the original summaries as training targets.

\textbf{Real data vs. Pseudo data} Table \ref{tab:train_mode_samsum_dialogsum_short_form} shows that training with pseudo data first and then with real data (\textbf{Pseudo+Real}) or training with only pseudo data (\textbf{Pseduo}) are the two best training modes. The former is better on SAMSum and the latter is better on DialogSum. Here we cannot say that real data is less effective because there is a huge difference in the size between real and pseudo data: real training data is only 1200 items on each dataset; while the generated pseudo data are 40451 and 35174 items on SAMSum and DialogSum, respectively. This on the one hand corroborates the effectiveness of the FEC approach based on data augmentation in the past, and on the other hand, implies that the combination of real and pseudo data is promising.

Regarding the performance on the form-based error categories: On both datasets, most of the edits are in the \textbf{Replacement} category (see Table \ref{tab:train_mode_samsum_tp_form} and Table \ref{tab:train_mode_dialogsum_tp_form} in Appendix \ref{sec:appendix_fig_tab}). Table \ref{tab:train_mode_samsum_dialogsum_short_form} illustrates that using the reference correction as the final training goal (\textbf{Real}, \textbf{Pseudo+Real}) performs poorly on the \textbf{Missing} category. This indicates that it is difficult for models to learn addition operations in manual correction. 

In addition, we also try to mix SAMSum and DialogSum as a corpus for constructing pseudo data. Table \ref{tab:corpus_samsum_dialogsum_form} in Appendix \ref{sec:appendix_fig_tab} illustrates that in some cases, the mixed construction has better results than the separate construction. For comparison, we still construct the pseudo data separately in the subsequent experiments.

\subsection{Performance across FEC Approaches}
Here we mainly show the results of data augmentation approaches on the category of content-based errors. Form-based results are shown in Table \ref{tab:approach_samsum_dialogsum_short_form} in Appendix \ref{sec:appendix_fig_tab}. Training modes are set to \textbf{Pseudo} and \textbf{Pseudo+Real}.

\textbf{Ent:ObjE} and \textbf{Pred:VerbE} are the two main error types (see Tables \ref{tab:approach_samsum_tp_content} and \ref{tab:approach_dialogsum_tp_content} in Appendix \ref{sec:appendix_fig_tab}), which coincide with our annotation results in Section \ref{sec:data_ann}. An important finding is that Tables \ref{tab:approach_samsum_short_content} (and Table \ref{tab:approach_dialogsum_short_content} in Appendix \ref{sec:appendix_fig_tab}) show that these methods based on data augmentation for generating corrected summaries directly show error-correcting power only for a few categories: \textbf{Ent:ObjE}, \textbf{Pred:VerbE}, \textbf{CorefE}, \textbf{NumE}, and \textbf{OthE}. We argue that this cannot be attributed only to the chance brought by the small percentage of some error categories. The strategy of data augmentation is an important factor. Because we notice the fact that the rule-based data augmentation approach performs swapping on numbers, and it has a relatively great performance on \textbf{NumE} on SAMSum, even though the percentage of \textbf{NumE} is small.

The infilling-based data augmentation method is generally inferior to the rule-based data augmentation method. Its performance also changes insignificantly after retraining. The particular structural information in the conversation summaries has to be further exploited. The infilling-based method sometimes performs better on \textbf{Pred:VerbE}. This may be due to the fact that it masks and predicts the relations in the reference summary when constructing pseudo data, with verb phrases in the relations.

In addition, both CCGS and Factpegasus perform poorly. Table \ref{tab:ccgs_samsum_dialogsum_short_form} illustrates that CCGS can only correct errors in the form of substitution. Table \ref{tab:factpegasus_samsum_dialogsum_short_form} illustrates that Factpegasus can only correct errors by deletion. This is consistent with their algorithms. Table \ref{tab:ccgs_samsum_dialogsum_tp_content} and Table \ref{tab:factpegasus_samsum_dialogsum_tp_content} in Appendix \ref{sec:appendix_fig_tab} illustrate that they can almost correct only one type of errors, \textbf{Ent:ObjE}.

However, the above findings would not have been available if we had used only factuality metrics (see Table \ref{tab:approach_samsum_fact}, Table \ref{tab:approach_dialogsum_fact}, Table \ref{tab:ccgs_samsum_dialogsum_fact} and Table \ref{tab:factpegasus_samsum_dialogsum_fact} in Appendix \ref{sec:appendix_fig_tab}). This illustrates the superiority of FERRANTI.

\section{Conclusion}
Our work establishes a new benchmark for model-agnostic factual error correction for dialogue summarization. Unlike previous studies, we manually correct factual errors in summaries. We point out the shortcomings of factuality metrics in FEC evaluation: They are not reliable enough and cannot provide more detailed information. For better evaluation, we propose FERRANTI, a reference-based evaluation framework and conduct thorough experiments on the performance of multiple FEC approaches under various settings. We have the following important findings: 

\begin{itemize}
  \item [1)]
Training FEC models with reference summaries from dialogue summarization datasets yields the best results of unreliable factuality metrics. There is an urgent need to change the evaluation methods for FEC models.
  \item [2)]
Introducing human-corrected summaries during the training of FEC models for dialogue summarization can improve their performance. Combining human-annotated data with synthetic data is a promising direction.
  \item [3)]
Current FEC models struggle to correct factual errors by addition and cannot address attribute errors, modality errors, link errors, etc.
\end{itemize}

For future work, it is feasible to apply FERRANTI to FEC for other summarization tasks.

\section*{Limitations}
Due to limited resources, the size of our annotated dataset is not large, with only 4000 items. In addition, we use an annotation paradigm where direct writing is the main focus with error labeling as a supplement. This is good for the coherence of the corrected summary and gives larger freedom to the annotator. In this case, it may be better to increase the number of reference corrections per sample. The datasets we select, SAMSum and DialogSum, are both short daily chat summarization datasets. For other domains or long dialogue summarization, our conclusion may not apply.

About FERRANTI, it can be continuously improved since we automatically classify and label factual errors for the first time.
It also relies on the lexical and syntactic nature of English.

\section*{Ethics Statement}
We recruit annotators through the campus BBS. They are completely free to decide whether to participate and can quit in the middle. They are paid \$15 per hour, more than the local minimum wage. No participants' personal information or payment information will be released. Some of the information is temporarily stored on the server and will be deleted at the end of the study.


The application of datasets, models, and tools in our study is consistent with their intended use and license. We hope the artifacts we release are to be used for academic research (non-commercial licence: CC BY-NC 4.0).


\section*{Acknowledgements}
This work was supported by National Key R\&D Program of China (2021YFF0901502), National Science Foundation of China (No. 62161160339), State Key Laboratory of Media Convergence Production Technology and Systems and Key Laboratory of Science, Technology and Standard in Press Industry (Key Laboratory of Intelligent Press Media Technology). We appreciate the anonymous reviewers for their helpful comments. Xiaojun Wan is the corresponding author.


\begin{thebibliography}{48}
\expandafter\ifx\csname natexlab\endcsname\relax\def\natexlab#1{#1}\fi

\bibitem[{Balachandran et~al.(2022)Balachandran, Hajishirzi, Cohen, and
  Tsvetkov}]{Balachandran2022CorrectingDF}
Vidhisha Balachandran, Hannaneh Hajishirzi, William Cohen, and Yulia Tsvetkov.
  2022.
\newblock \href {https://arxiv.org/abs/2210.12378} {Correcting diverse factual
  errors in abstractive summarization via post-editing and language model
  infilling}.
\newblock \emph{Computing Research Repository}, arXiv:2210.12378.

\bibitem[{Bryant et~al.(2017)Bryant, Felice, and
  Briscoe}]{bryant-etal-2017-automatic}
Christopher Bryant, Mariano Felice, and Ted Briscoe. 2017.
\newblock \href {https://doi.org/10.18653/v1/P17-1074} {Automatic annotation
  and evaluation of error types for grammatical error correction}.
\newblock In \emph{Proceedings of the 55th Annual Meeting of the Association
  for Computational Linguistics (Volume 1: Long Papers)}, pages 793--805,
  Vancouver, Canada. Association for Computational Linguistics.

\bibitem[{Cao et~al.(2020)Cao, Dong, Wu, and Cheung}]{cao-etal-2020-factual}
Meng Cao, Yue Dong, Jiapeng Wu, and Jackie Chi~Kit Cheung. 2020.
\newblock \href {https://doi.org/10.18653/v1/2020.emnlp-main.506} {Factual
  error correction for abstractive summarization models}.
\newblock In \emph{Proceedings of the 2020 Conference on Empirical Methods in
  Natural Language Processing (EMNLP)}, pages 6251--6258, Online. Association
  for Computational Linguistics.

\bibitem[{Chen and Yang(2020)}]{chen-yang-2020-multi}
Jiaao Chen and Diyi Yang. 2020.
\newblock \href {https://doi.org/10.18653/v1/2020.emnlp-main.336} {Multi-view
  sequence-to-sequence models with conversational structure for abstractive
  dialogue summarization}.
\newblock In \emph{Proceedings of the 2020 Conference on Empirical Methods in
  Natural Language Processing (EMNLP)}, pages 4106--4118, Online. Association
  for Computational Linguistics.

\bibitem[{Chen and Yang(2021)}]{chen-yang-2021-structure}
Jiaao Chen and Diyi Yang. 2021.
\newblock \href {https://doi.org/10.18653/v1/2021.naacl-main.109}
  {Structure-aware abstractive conversation summarization via discourse and
  action graphs}.
\newblock In \emph{Proceedings of the 2021 Conference of the North American
  Chapter of the Association for Computational Linguistics: Human Language
  Technologies}, pages 1380--1391, Online. Association for Computational
  Linguistics.

\bibitem[{Chen et~al.(2022)Chen, Xu, Zeng, Sun, Li, and
  Xiao}]{Chen2022ConvergeTT}
Jiangjie Chen, Rui Xu, Wenyuan Zeng, Changzhi Sun, Lei Li, and Yanghua Xiao.
  2022.
\newblock \href {https://arxiv.org/abs/2211.12130} {Converge to the truth:
  Factual error correction via iterative constrained editing}.
\newblock \emph{Computing Research Repository}, arXiv:2211.12130.

\bibitem[{Chen et~al.(2021{\natexlab{a}})Chen, Zhang, Sone, and
  Roth}]{chen-etal-2021-improving}
Sihao Chen, Fan Zhang, Kazoo Sone, and Dan Roth. 2021{\natexlab{a}}.
\newblock \href {https://doi.org/10.18653/v1/2021.naacl-main.475} {Improving
  faithfulness in abstractive summarization with contrast candidate generation
  and selection}.
\newblock In \emph{Proceedings of the 2021 Conference of the North American
  Chapter of the Association for Computational Linguistics: Human Language
  Technologies}, pages 5935--5941, Online. Association for Computational
  Linguistics.

\bibitem[{Chen et~al.(2021{\natexlab{b}})Chen, Liu, Chen, and
  Zhang}]{chen-etal-2021-dialogsum}
Yulong Chen, Yang Liu, Liang Chen, and Yue Zhang. 2021{\natexlab{b}}.
\newblock \href {https://doi.org/10.18653/v1/2021.findings-acl.449}
  {{D}ialog{S}um: {A} real-life scenario dialogue summarization dataset}.
\newblock In \emph{Findings of the Association for Computational Linguistics:
  ACL-IJCNLP 2021}, pages 5062--5074, Online. Association for Computational
  Linguistics.

\bibitem[{Chollampatt et~al.(2020)Chollampatt, Susanto, Tan, and
  Szymanska}]{chollampatt-etal-2020-automatic}
Shamil Chollampatt, Raymond~Hendy Susanto, Liling Tan, and Ewa Szymanska. 2020.
\newblock \href {https://doi.org/10.18653/v1/2020.emnlp-main.217} {Can
  automatic post-editing improve {NMT}?}
\newblock In \emph{Proceedings of the 2020 Conference on Empirical Methods in
  Natural Language Processing (EMNLP)}, pages 2736--2746, Online. Association
  for Computational Linguistics.

\bibitem[{Dahlmeier and Ng(2012)}]{dahlmeier-ng-2012-better}
Daniel Dahlmeier and Hwee~Tou Ng. 2012.
\newblock \href {https://aclanthology.org/N12-1067} {Better evaluation for
  grammatical error correction}.
\newblock In \emph{Proceedings of the 2012 Conference of the North {A}merican
  Chapter of the Association for Computational Linguistics: Human Language
  Technologies}, pages 568--572, Montr{\'e}al, Canada. Association for
  Computational Linguistics.

\bibitem[{Devaraj et~al.(2022)Devaraj, Sheffield, Wallace, and
  Li}]{devaraj-etal-2022-evaluating}
Ashwin Devaraj, William Sheffield, Byron Wallace, and Junyi~Jessy Li. 2022.
\newblock \href {https://doi.org/10.18653/v1/2022.acl-long.506} {Evaluating
  factuality in text simplification}.
\newblock In \emph{Proceedings of the 60th Annual Meeting of the Association
  for Computational Linguistics (Volume 1: Long Papers)}, pages 7331--7345,
  Dublin, Ireland. Association for Computational Linguistics.

\bibitem[{Devlin et~al.(2019)Devlin, Chang, Lee, and
  Toutanova}]{devlin-etal-2019-bert}
Jacob Devlin, Ming-Wei Chang, Kenton Lee, and Kristina Toutanova. 2019.
\newblock \href {https://doi.org/10.18653/v1/N19-1423} {{BERT}: Pre-training of
  deep bidirectional transformers for language understanding}.
\newblock In \emph{Proceedings of the 2019 Conference of the North {A}merican
  Chapter of the Association for Computational Linguistics: Human Language
  Technologies, Volume 1 (Long and Short Papers)}, pages 4171--4186,
  Minneapolis, Minnesota. Association for Computational Linguistics.

\bibitem[{Dong et~al.(2019)Dong, Yang, Wang, Wei, Liu, Wang, Gao, Zhou, and
  Hon}]{NEURIPS2019_c20bb2d9}
Li~Dong, Nan Yang, Wenhui Wang, Furu Wei, Xiaodong Liu, Yu~Wang, Jianfeng Gao,
  Ming Zhou, and Hsiao-Wuen Hon. 2019.
\newblock \href
  {https://proceedings.neurips.cc/paper/2019/file/c20bb2d9a50d5ac1f713f8b34d9aac5a-Paper.pdf}
  {Unified language model pre-training for natural language understanding and
  generation}.
\newblock In \emph{Advances in Neural Information Processing Systems},
  volume~32. Curran Associates, Inc.

\bibitem[{Dong et~al.(2020)Dong, Wang, Gan, Cheng, Cheung, and
  Liu}]{dong-etal-2020-multi}
Yue Dong, Shuohang Wang, Zhe Gan, Yu~Cheng, Jackie Chi~Kit Cheung, and Jingjing
  Liu. 2020.
\newblock \href {https://doi.org/10.18653/v1/2020.emnlp-main.749} {Multi-fact
  correction in abstractive text summarization}.
\newblock In \emph{Proceedings of the 2020 Conference on Empirical Methods in
  Natural Language Processing (EMNLP)}, pages 9320--9331, Online. Association
  for Computational Linguistics.

\bibitem[{Durmus et~al.(2020)Durmus, He, and Diab}]{durmus-etal-2020-feqa}
Esin Durmus, He~He, and Mona Diab. 2020.
\newblock \href {https://doi.org/10.18653/v1/2020.acl-main.454} {{FEQA}: A
  question answering evaluation framework for faithfulness assessment in
  abstractive summarization}.
\newblock In \emph{Proceedings of the 58th Annual Meeting of the Association
  for Computational Linguistics}, pages 5055--5070, Online. Association for
  Computational Linguistics.

\bibitem[{Fabbri et~al.(2021{\natexlab{a}})Fabbri, Rahman, Rizvi, Wang, Li,
  Mehdad, and Radev}]{fabbri-etal-2021-convosumm}
Alexander Fabbri, Faiaz Rahman, Imad Rizvi, Borui Wang, Haoran Li, Yashar
  Mehdad, and Dragomir Radev. 2021{\natexlab{a}}.
\newblock \href {https://doi.org/10.18653/v1/2021.acl-long.535} {{C}onvo{S}umm:
  Conversation summarization benchmark and improved abstractive summarization
  with argument mining}.
\newblock In \emph{Proceedings of the 59th Annual Meeting of the Association
  for Computational Linguistics and the 11th International Joint Conference on
  Natural Language Processing (Volume 1: Long Papers)}, pages 6866--6880,
  Online. Association for Computational Linguistics.

\bibitem[{Fabbri et~al.(2022{\natexlab{a}})Fabbri, Wu, Liu, and
  Xiong}]{fabbri-etal-2022-qafacteval}
Alexander Fabbri, Chien-Sheng Wu, Wenhao Liu, and Caiming Xiong.
  2022{\natexlab{a}}.
\newblock \href {https://doi.org/10.18653/v1/2022.naacl-main.187}
  {{QAF}act{E}val: Improved {QA}-based factual consistency evaluation for
  summarization}.
\newblock In \emph{Proceedings of the 2022 Conference of the North American
  Chapter of the Association for Computational Linguistics: Human Language
  Technologies}, pages 2587--2601, Seattle, United States. Association for
  Computational Linguistics.

\bibitem[{Fabbri et~al.(2022{\natexlab{b}})Fabbri, Choubey, Vig, Wu, and
  Xiong}]{Fabbri2022ImprovingFC}
Alexander~R. Fabbri, Prafulla~Kumar Choubey, Jesse Vig, Chien-Sheng Wu, and
  Caiming Xiong. 2022{\natexlab{b}}.
\newblock \href {https://arxiv.org/abs/2211.06196} {Improving factual
  consistency in summarization with compression-based post-editing}.
\newblock \emph{Computing Research Repository}, abs/2211.06196.

\bibitem[{Fabbri et~al.(2021{\natexlab{b}})Fabbri, Kry{\'s}ci{\'n}ski, McCann,
  Xiong, Socher, and Radev}]{fabbri-etal-2021-summeval}
Alexander~R. Fabbri, Wojciech Kry{\'s}ci{\'n}ski, Bryan McCann, Caiming Xiong,
  Richard Socher, and Dragomir Radev. 2021{\natexlab{b}}.
\newblock \href {https://doi.org/10.1162/tacl_a_00373} {{S}umm{E}val:
  Re-evaluating summarization evaluation}.
\newblock \emph{Transactions of the Association for Computational Linguistics},
  9:391--409.

\bibitem[{Falke et~al.(2019)Falke, Ribeiro, Utama, Dagan, and
  Gurevych}]{falke-etal-2019-ranking}
Tobias Falke, Leonardo F.~R. Ribeiro, Prasetya~Ajie Utama, Ido Dagan, and Iryna
  Gurevych. 2019.
\newblock \href {https://doi.org/10.18653/v1/P19-1213} {Ranking generated
  summaries by correctness: An interesting but challenging application for
  natural language inference}.
\newblock In \emph{Proceedings of the 57th Annual Meeting of the Association
  for Computational Linguistics}, pages 2214--2220, Florence, Italy.
  Association for Computational Linguistics.

\bibitem[{Felice et~al.(2016)Felice, Bryant, and
  Briscoe}]{felice-etal-2016-automatic}
Mariano Felice, Christopher Bryant, and Ted Briscoe. 2016.
\newblock \href {https://aclanthology.org/C16-1079} {Automatic extraction of
  learner errors in {ESL} sentences using linguistically enhanced alignments}.
\newblock In \emph{Proceedings of {COLING} 2016, the 26th International
  Conference on Computational Linguistics: Technical Papers}, pages 825--835,
  Osaka, Japan. The COLING 2016 Organizing Committee.

\bibitem[{Gao and Wan(2022)}]{gao-wan-2022-dialsummeval}
Mingqi Gao and Xiaojun Wan. 2022.
\newblock \href {https://doi.org/10.18653/v1/2022.naacl-main.418}
  {{D}ial{S}umm{E}val: Revisiting summarization evaluation for dialogues}.
\newblock In \emph{Proceedings of the 2022 Conference of the North American
  Chapter of the Association for Computational Linguistics: Human Language
  Technologies}, pages 5693--5709, Seattle, United States. Association for
  Computational Linguistics.

\bibitem[{Gliwa et~al.(2019)Gliwa, Mochol, Biesek, and
  Wawer}]{gliwa-etal-2019-samsum}
Bogdan Gliwa, Iwona Mochol, Maciej Biesek, and Aleksander Wawer. 2019.
\newblock \href {https://doi.org/10.18653/v1/D19-5409} {{SAMS}um corpus: A
  human-annotated dialogue dataset for abstractive summarization}.
\newblock In \emph{Proceedings of the 2nd Workshop on New Frontiers in
  Summarization}, pages 70--79, Hong Kong, China. Association for Computational
  Linguistics.

\bibitem[{Goyal and Durrett(2020)}]{goyal-durrett-2020-evaluating}
Tanya Goyal and Greg Durrett. 2020.
\newblock \href {https://doi.org/10.18653/v1/2020.findings-emnlp.322}
  {Evaluating factuality in generation with dependency-level entailment}.
\newblock In \emph{Findings of the Association for Computational Linguistics:
  EMNLP 2020}, pages 3592--3603, Online. Association for Computational
  Linguistics.

\bibitem[{Goyal and Durrett(2021)}]{goyal-durrett-2021-annotating}
Tanya Goyal and Greg Durrett. 2021.
\newblock \href {https://doi.org/10.18653/v1/2021.naacl-main.114} {Annotating
  and modeling fine-grained factuality in summarization}.
\newblock In \emph{Proceedings of the 2021 Conference of the North American
  Chapter of the Association for Computational Linguistics: Human Language
  Technologies}, pages 1449--1462, Online. Association for Computational
  Linguistics.

\bibitem[{Kryscinski et~al.(2020)Kryscinski, McCann, Xiong, and
  Socher}]{kryscinski-etal-2020-evaluating}
Wojciech Kryscinski, Bryan McCann, Caiming Xiong, and Richard Socher. 2020.
\newblock \href {https://doi.org/10.18653/v1/2020.emnlp-main.750} {Evaluating
  the factual consistency of abstractive text summarization}.
\newblock In \emph{Proceedings of the 2020 Conference on Empirical Methods in
  Natural Language Processing (EMNLP)}, pages 9332--9346, Online. Association
  for Computational Linguistics.

\bibitem[{Laban et~al.(2022)Laban, Schnabel, Bennett, and
  Hearst}]{laban-etal-2022-summac}
Philippe Laban, Tobias Schnabel, Paul~N. Bennett, and Marti~A. Hearst. 2022.
\newblock \href {https://doi.org/10.1162/tacl_a_00453} {{S}umma{C}: Re-visiting
  {NLI}-based models for inconsistency detection in summarization}.
\newblock \emph{Transactions of the Association for Computational Linguistics},
  10:163--177.

\bibitem[{Lewis et~al.(2020)Lewis, Liu, Goyal, Ghazvininejad, Mohamed, Levy,
  Stoyanov, and Zettlemoyer}]{lewis-etal-2020-bart}
Mike Lewis, Yinhan Liu, Naman Goyal, Marjan Ghazvininejad, Abdelrahman Mohamed,
  Omer Levy, Veselin Stoyanov, and Luke Zettlemoyer. 2020.
\newblock \href {https://doi.org/10.18653/v1/2020.acl-main.703} {{BART}:
  Denoising sequence-to-sequence pre-training for natural language generation,
  translation, and comprehension}.
\newblock In \emph{Proceedings of the 58th Annual Meeting of the Association
  for Computational Linguistics}, pages 7871--7880, Online. Association for
  Computational Linguistics.

\bibitem[{Liu et~al.(2019)Liu, Ott, Goyal, Du, Joshi, Chen, Levy, Lewis,
  Zettlemoyer, and Stoyanov}]{liu2019roberta}
Yinhan Liu, Myle Ott, Naman Goyal, Jingfei Du, Mandar Joshi, Danqi Chen, Omer
  Levy, Mike Lewis, Luke Zettlemoyer, and Veselin Stoyanov. 2019.
\newblock \href {https://arxiv.org/abs/1907.11692} {Roberta: {A} robustly
  optimized {BERT} pretraining approach}.
\newblock \emph{Computing Research Repository}, arXiv:1907.11692.

\bibitem[{Liu and Chen(2021)}]{liu-chen-2021-controllable}
Zhengyuan Liu and Nancy Chen. 2021.
\newblock \href {https://doi.org/10.18653/v1/2021.emnlp-main.8} {Controllable
  neural dialogue summarization with personal named entity planning}.
\newblock In \emph{Proceedings of the 2021 Conference on Empirical Methods in
  Natural Language Processing}, pages 92--106, Online and Punta Cana, Dominican
  Republic. Association for Computational Linguistics.

\bibitem[{Liu et~al.(2021)Liu, Shi, and Chen}]{liu-etal-2021-coreference}
Zhengyuan Liu, Ke~Shi, and Nancy Chen. 2021.
\newblock \href {https://aclanthology.org/2021.sigdial-1.53} {Coreference-aware
  dialogue summarization}.
\newblock In \emph{Proceedings of the 22nd Annual Meeting of the Special
  Interest Group on Discourse and Dialogue}, pages 509--519, Singapore and
  Online. Association for Computational Linguistics.

\bibitem[{Maynez et~al.(2020)Maynez, Narayan, Bohnet, and
  McDonald}]{maynez-etal-2020-faithfulness}
Joshua Maynez, Shashi Narayan, Bernd Bohnet, and Ryan McDonald. 2020.
\newblock \href {https://doi.org/10.18653/v1/2020.acl-main.173} {On
  faithfulness and factuality in abstractive summarization}.
\newblock In \emph{Proceedings of the 58th Annual Meeting of the Association
  for Computational Linguistics}, pages 1906--1919, Online. Association for
  Computational Linguistics.

\bibitem[{Pagnoni et~al.(2021)Pagnoni, Balachandran, and
  Tsvetkov}]{pagnoni-etal-2021-understanding}
Artidoro Pagnoni, Vidhisha Balachandran, and Yulia Tsvetkov. 2021.
\newblock \href {https://doi.org/10.18653/v1/2021.naacl-main.383}
  {Understanding factuality in abstractive summarization with {FRANK}: A
  benchmark for factuality metrics}.
\newblock In \emph{Proceedings of the 2021 Conference of the North American
  Chapter of the Association for Computational Linguistics: Human Language
  Technologies}, pages 4812--4829, Online. Association for Computational
  Linguistics.

\bibitem[{Popovi{\'c}(2015)}]{popovic-2015-chrf}
Maja Popovi{\'c}. 2015.
\newblock \href {https://doi.org/10.18653/v1/W15-3049} {chr{F}: character
  n-gram {F}-score for automatic {MT} evaluation}.
\newblock In \emph{Proceedings of the Tenth Workshop on Statistical Machine
  Translation}, pages 392--395, Lisbon, Portugal. Association for Computational
  Linguistics.

\bibitem[{Raffel et~al.(2022)Raffel, Shazeer, Roberts, Lee, Narang, Matena,
  Zhou, Li, and Liu}]{raffel2020exploring}
Colin Raffel, Noam Shazeer, Adam Roberts, Katherine Lee, Sharan Narang, Michael
  Matena, Yanqi Zhou, Wei Li, and Peter~J. Liu. 2022.
\newblock Exploring the limits of transfer learning with a unified text-to-text
  transformer.
\newblock \emph{J. Mach. Learn. Res.}, 21(1).

\bibitem[{Scialom et~al.(2021)Scialom, Dray, Lamprier, Piwowarski, Staiano,
  Wang, and Gallinari}]{scialom-etal-2021-questeval}
Thomas Scialom, Paul-Alexis Dray, Sylvain Lamprier, Benjamin Piwowarski, Jacopo
  Staiano, Alex Wang, and Patrick Gallinari. 2021.
\newblock \href {https://doi.org/10.18653/v1/2021.emnlp-main.529}
  {{Q}uest{E}val: Summarization asks for fact-based evaluation}.
\newblock In \emph{Proceedings of the 2021 Conference on Empirical Methods in
  Natural Language Processing}, pages 6594--6604, Online and Punta Cana,
  Dominican Republic. Association for Computational Linguistics.

\bibitem[{Scialom et~al.(2019)Scialom, Lamprier, Piwowarski, and
  Staiano}]{scialom-etal-2019-answers}
Thomas Scialom, Sylvain Lamprier, Benjamin Piwowarski, and Jacopo Staiano.
  2019.
\newblock \href {https://doi.org/10.18653/v1/D19-1320} {Answers unite!
  unsupervised metrics for reinforced summarization models}.
\newblock In \emph{Proceedings of the 2019 Conference on Empirical Methods in
  Natural Language Processing and the 9th International Joint Conference on
  Natural Language Processing (EMNLP-IJCNLP)}, pages 3246--3256, Hong Kong,
  China. Association for Computational Linguistics.

\bibitem[{Shah et~al.(2020)Shah, Schuster, and
  Barzilay}]{Shah_Schuster_Barzilay_2020}
Darsh Shah, Tal Schuster, and Regina Barzilay. 2020.
\newblock \href {https://doi.org/10.1609/aaai.v34i05.6406} {Automatic
  fact-guided sentence modification}.
\newblock \emph{Proceedings of the AAAI Conference on Artificial Intelligence},
  34(05):8791--8798.

\bibitem[{Snover et~al.(2006)Snover, Dorr, Shwartz, Micciulla, and
  Makhoul}]{snover2006ter}
Matthew Snover, Bonnie Dorr, Richard Shwartz, Linnea Micciulla, and John
  Makhoul. 2006.
\newblock \href {http://www.mt-archive.info/05/AMTA-2006-Snover.pdf} {A study
  of translation edit rate with targeted human annotation}.
\newblock In \emph{Proceedings of the Seventh Conference of the Association for
  Machine Translation in the Americas}.

\bibitem[{Tang et~al.(2022)Tang, Nair, Wang, Wang, Desai, Wade, Li,
  Celikyilmaz, Mehdad, and Radev}]{tang-etal-2022-confit}
Xiangru Tang, Arjun Nair, Borui Wang, Bingyao Wang, Jai Desai, Aaron Wade,
  Haoran Li, Asli Celikyilmaz, Yashar Mehdad, and Dragomir Radev. 2022.
\newblock \href {https://doi.org/10.18653/v1/2022.naacl-main.415} {{CONFIT}:
  Toward faithful dialogue summarization with linguistically-informed
  contrastive fine-tuning}.
\newblock In \emph{Proceedings of the 2022 Conference of the North American
  Chapter of the Association for Computational Linguistics: Human Language
  Technologies}, pages 5657--5668, Seattle, United States. Association for
  Computational Linguistics.

\bibitem[{Thorne and Vlachos(2021)}]{thorne-vlachos-2021-evidence}
James Thorne and Andreas Vlachos. 2021.
\newblock \href {https://doi.org/10.18653/v1/2021.acl-long.256} {Evidence-based
  factual error correction}.
\newblock In \emph{Proceedings of the 59th Annual Meeting of the Association
  for Computational Linguistics and the 11th International Joint Conference on
  Natural Language Processing (Volume 1: Long Papers)}, pages 3298--3309,
  Online. Association for Computational Linguistics.

\bibitem[{Wan and Bansal(2022)}]{wan-bansal-2022-factpegasus}
David Wan and Mohit Bansal. 2022.
\newblock \href {https://doi.org/10.18653/v1/2022.naacl-main.74}
  {{F}act{PEGASUS}: Factuality-aware pre-training and fine-tuning for
  abstractive summarization}.
\newblock In \emph{Proceedings of the 2022 Conference of the North American
  Chapter of the Association for Computational Linguistics: Human Language
  Technologies}, pages 1010--1028, Seattle, United States. Association for
  Computational Linguistics.

\bibitem[{Wang et~al.(2022)Wang, Zhang, Zhang, Chen, and
  Li}]{wang2022analyzing}
Bin Wang, Chen Zhang, Yan Zhang, Yiming Chen, and Haizhou Li. 2022.
\newblock \href {https://arxiv.org/abs/2210.11777} {Analyzing and evaluating
  faithfulness in dialogue summarization}.
\newblock \emph{Computing Research Repository}, arXiv:2210.11777.

\bibitem[{Wu et~al.(2021)Wu, Liu, Liu, Stenetorp, and
  Xiong}]{wu-etal-2021-controllable}
Chien-Sheng Wu, Linqing Liu, Wenhao Liu, Pontus Stenetorp, and Caiming Xiong.
  2021.
\newblock \href {https://doi.org/10.18653/v1/2021.findings-acl.454}
  {Controllable abstractive dialogue summarization with sketch supervision}.
\newblock In \emph{Findings of the Association for Computational Linguistics:
  ACL-IJCNLP 2021}, pages 5108--5122, Online. Association for Computational
  Linguistics.

\bibitem[{Xu et~al.(2016)Xu, Napoles, Pavlick, Chen, and
  Callison-Burch}]{xu-etal-2016-optimizing}
Wei Xu, Courtney Napoles, Ellie Pavlick, Quanze Chen, and Chris Callison-Burch.
  2016.
\newblock \href {https://doi.org/10.1162/tacl_a_00107} {Optimizing statistical
  machine translation for text simplification}.
\newblock \emph{Transactions of the Association for Computational Linguistics},
  4:401--415.

\bibitem[{Yuan et~al.(2021)Yuan, Neubig, and Liu}]{yuan2021bartscore}
Weizhe Yuan, Graham Neubig, and Pengfei Liu. 2021.
\newblock \href {https://arxiv.org/abs/2106.11520} {Bartscore: Evaluating
  generated text as text generation}.
\newblock \emph{Computing Research Repository}, arXiv:2106.11520.
\newblock Version 2.

\bibitem[{Zhang et~al.(2020)Zhang, Zhao, Saleh, and Liu}]{zhang2020pegasus}
Jingqing Zhang, Yao Zhao, Mohammad Saleh, and Peter Liu. 2020.
\newblock Pegasus: Pre-training with extracted gap-sentences for abstractive
  summarization.
\newblock In \emph{International Conference on Machine Learning}, pages
  11328--11339. PMLR.

\bibitem[{Zhu et~al.(2021)Zhu, Hinthorn, Xu, Zeng, Zeng, Huang, and
  Jiang}]{zhu-etal-2021-enhancing}
Chenguang Zhu, William Hinthorn, Ruochen Xu, Qingkai Zeng, Michael Zeng,
  Xuedong Huang, and Meng Jiang. 2021.
\newblock \href {https://doi.org/10.18653/v1/2021.naacl-main.58} {Enhancing
  factual consistency of abstractive summarization}.
\newblock In \emph{Proceedings of the 2021 Conference of the North American
  Chapter of the Association for Computational Linguistics: Human Language
  Technologies}, pages 718--733, Online. Association for Computational
  Linguistics.

\end{thebibliography}
\bibliographystyle{acl_natbib}

\appendix

\section{Details of Annotation}
\label{sec:appdix_example}

The annotators were told that the collected data would be used for academic study. In total, 10 people participated in the annotation. Two people read the annotation guidelines and then abandoned further annotation. One person annotated the small part used for testing and then gave up on further annotation. The other seven qualified participants who continued to annotate are from Asia. Three of them are female and four of them are male. One annotated three batches, another annotated two batches, and the others annotated one batch each. The screenshot of the annotation interface is shown in Figure \ref{fig:interface} in Appendix \ref{sec:appendix_fig_tab}. Considering the space for corrected summaries is relatively narrow, we provide an excel file for annotators to help them write the corrected summaries (shown in Figure \ref{fig:interface_excel} in Appendix \ref{sec:appendix_fig_tab}). They can copy what they write in the excel file and paste it into the interface. They decide whether to use the excel file according to their needs. We use what they submit in the interface as the final result.

We provide the same definition of error categories for annotators as \citet{pagnoni-etal-2021-understanding}, but with different examples because the original examples are news summaries. They are shown in Table \ref{tab:EntE}, Table \ref{tab:PredE}, Table \ref{tab:CircE}, Table \ref{tab:CorefE}, Table \ref{tab:LinkE}, Table \ref{tab:OutE}, and Table \ref{tab:GramE} in Appendix \ref{sec:appdix_example}.

\begin{table}[h]
\begin{tabular}{|c|}
\hline
\textbf{Entity Error (EntE)} \\ \hline
\textbf{Dialogue} \\ \hline
\multicolumn{1}{|l|}{\begin{tabular}[c]{@{}l@{}} \textbf{Ola}: Hey running late    \\ \textbf{Ola}: I should be free by 8    \\ 
\textbf{Kurt}: Sure no prob, call me\end{tabular}} \\ \hline
\textbf{Original   Summary} \\ \hline
\multicolumn{1}{|l|}{Ola will be late. \textcolor{red}{Kurt} will call him by   8.} \\ \hline
\textbf{Corrected   Summary} \\ \hline
\multicolumn{1}{|l|}{Ola will be late. He will call Kurt.} \\ \hline
\end{tabular}
\caption{An example of Entity Error.}
\label{tab:EntE}
\end{table}

\begin{table}[h]
\begin{tabular}{|c|}
\hline
\textbf{Coreference Error (CorefE)} \\ \hline
\textbf{Dialogue} \\ \hline
\multicolumn{1}{|l|}{\begin{tabular}[c]{@{}l@{}} \textbf{Ola}: Hey running late    \\ \textbf{Ola}: I should be free by 8    \\ 
\textbf{Kurt}: Sure no prob, call me\end{tabular}} \\ \hline
\textbf{Original   Summary} \\ \hline
\multicolumn{1}{|l|}{Ola will be late. Kurt will call \textcolor{red}{him} by   8.} \\ \hline
\textbf{Corrected   Summary} \\ \hline
\multicolumn{1}{|l|}{Ola will be late. He will call Kurt.} \\ \hline
\end{tabular}
\caption{An example of Coreference Error.}
\label{tab:CorefE}
\end{table}

\begin{table}[h]
\resizebox{\linewidth}{!}{
\begin{tabular}{|c|}
\hline
\textbf{Out   of Article Error (OutE)} \\ \hline
\textbf{Dialogue} \\ \hline
\multicolumn{1}{|l|}{\begin{tabular}[c]{@{}l@{}}\textbf{Dave}: Hey, is Nicky still at your place?   Her phone is off\\ \textbf{Sam}: She just left\\ \textbf{Dave}: Thanks!\end{tabular}} \\ \hline
\textbf{Original   Summary} \\ \hline
\multicolumn{1}{|l|}{Nicky just \textcolor{red}{left her phone} at Dave’s place   .} \\ \hline
\textbf{Corrected   Summary} \\ \hline
\multicolumn{1}{|l|}{Nicky just left Dave’s place .} \\ \hline
\end{tabular}
}
\caption{An example of Out of Article Error.}
\label{tab:OutE}
\end{table}

\begin{table*}[]
\resizebox{\linewidth}{!}{
\begin{tabular}{|c|}
\hline
\textbf{Predicate Error (PredE)} \\ \hline
\textbf{Dialogue} \\ \hline
\multicolumn{1}{|l|}{\begin{tabular}[c]{@{}l@{}}\textbf{Will}: hey babe, what do you want for dinner tonight?\\ \textbf{Emma}: gah, don't even worry about it tonight\\ \textbf{Will}: what do you mean? everything ok?\\ \textbf{Emma}: not really, but it's ok, don't worry about cooking though, I'm not hungry\\ \textbf{Will}: Well what time will you be home?\\ \textbf{Emma}: soon, hopefully\\ \textbf{Will}: you sure? Maybe you want me to pick you up?\\ \textbf{Emma}: no no it's alright. I'll be home soon, i'll tell you when I get home. \\ \textbf{Will}: Alright, love you. \\ \textbf{Emma}: love you too.\end{tabular}} \\ \hline
\textbf{Original   Summary} \\ \hline
\multicolumn{1}{|l|}{Emma \textcolor{red}{doesn't want to cook dinner} tonight. She will tell Will when she gets home.} \\ \hline
\textbf{Corrected   Summary} \\ \hline
\multicolumn{1}{|l|}{Emma is not hungry tonight. She will tell Will when she gets home.} \\ \hline
\end{tabular}
}
\caption{An example of Predicate Error.}
\label{tab:PredE}
\end{table*}

\begin{table*}[]
\resizebox{\linewidth}{!}{
\begin{tabular}{|c|}
\hline
\textbf{Circumstance Error (CircE)} \\ \hline
\textbf{Dialogue} \\ \hline
\multicolumn{1}{|l|}{\begin{tabular}[c]{@{}l@{}}\textbf{Lenny}: Babe, can you help me with   something?    \\ \textbf{Bob}: Sure, what's up? \\ \textbf{Lenny}: Which one should I pick? \\ \textbf{Bob}: Send me photos \\ \textbf{Lenny}: \textless{}file\_photo\textgreater \\ \textbf{Lenny}: \textless{}file\_photo\textgreater \\ \textbf{Lenny}: \textless{}file\_photo\textgreater \\ \textbf{Bob}: I like the first ones best \\ \textbf{Lenny}: But I already have purple trousers. Does it make sense to have two pairs? \\ \textbf{Bob}: I have four black pairs :D :D \\ \textbf{Lenny}: yeah, but shouldn't I pick a different color? \\ \textbf{Bob}: what matters is what you'll give you the most outfit options \\ \textbf{Lenny}: So I guess I'll buy the first or the third pair then \\ \textbf{Bob}: Pick the best quality then \\ \textbf{Lenny}: ur right, thx | \\ \textbf{Bob}: no prob :)\end{tabular}} \\ \hline
\textbf{Original   Summary} \\ \hline
\multicolumn{1}{|l|}{Lenny will buy the first or the third pair of purple trousers \textcolor{red}{for Bob.}} \\ \hline
\textbf{Corrected   Summary} \\ \hline
\multicolumn{1}{|l|}{Lenny will buy the first or the third pair of purple trousers.} \\ \hline
\end{tabular}
}
\caption{An example of Circumstance Error.}
\label{tab:CircE}
\end{table*}

\begin{table*}[h]
\resizebox{\linewidth}{!}{
\begin{tabular}{|c|}
\hline
\textbf{Discourse   Link Error (LinkE)} \\ \hline
\textbf{Dialogue} \\ \hline
\multicolumn{1}{|l|}{\begin{tabular}[c]{@{}l@{}}The first vaccine for Ebola was approved   by the FDA in 2019 in the US, five years after the initial outbreak in 2014.\\ To produce the   vaccine, scientists had to sequence the DNA of Ebola, then identify possible vaccines, and finally\\ show successful clinical trials. Scientists say a vaccine for COVID-19 is unlikely to be   ready this year, although\\ clinical trials have already started.\end{tabular}} \\ \hline
\textbf{Original   Summary} \\ \hline
\multicolumn{1}{|l|}{To produce the vaccine, scientists have   to show successful human trials, \textcolor{red}{then} sequence the DNA of the virus.} \\ \hline
\textbf{Corrected   Summary} \\ \hline
\multicolumn{1}{|l|}{To produce the vaccine, scientists have   to show successful human trials, after sequence the DNA of the virus.} \\ \hline
\end{tabular}
}
\caption{An example of Discourse Link Error. This example is taken from \citet{pagnoni-etal-2021-understanding}, and we add a corrected summary.}
\label{tab:LinkE}
\end{table*}

\begin{table*}[]
\resizebox{\linewidth}{!}{
\begin{tabular}{|c|}
\hline
\textbf{Grammatical   Error (GramE)} \\ \hline
\textbf{Dialogue} \\ \hline
\multicolumn{1}{|l|}{\begin{tabular}[c]{@{}l@{}}\textbf{Everett}: Ralph asked me if i could give him your phone number, is that cool?  \\ \textbf{Amy}: who's ralph? \\ \textbf{Everett}: my friend, i introduced him to you at the pub last week, tall, brown hair, weird laugh... \\ \textbf{Amy}: oh i remember him now, is he a psycho? \\ \textbf{Everett}: no \\ \textbf{Amy}: ok, he can have my number\end{tabular}} \\ \hline
\textbf{Original   Summary} \\ \hline
\multicolumn{1}{|l|}{Everett will give \textcolor{red}{him him} phone number .} \\ \hline
\textbf{Corrected   Summary} \\ \hline
\multicolumn{1}{|l|}{Everett will give   Ralph Amy’s phone number .} \\ \hline
\end{tabular}
}
\caption{An example of Grammatical Error.}
\label{tab:GramE}
\end{table*}

\section{Details of the use of factuality metrics}

For \textbf{FactCC}\footnote{  \url{https://github.com/salesforce/factCC}} and \textbf{DAE} \footnote{ \url{https://github.com/tagoyal/factuality-datasets}}, We follow the way \citet{pagnoni-etal-2021-understanding} used it. The summary is split into sentences by NLTK \footnote{version 3.7, \url{https://www.nltk.org/}}. Each sentence is classified as \texttt{CORRECT} or \texttt{INCORRECT}. The factual score of a summary is represented as the ratio of factually correct sentences. 

For \textbf{QuestEval} \footnote{  \url{https://github.com/ThomasScialom/QuestEval}}, we use the reference-less mode. For \textbf{BARTScore} \footnote{  \url{https://github.com/neulab/BARTScore}}, we use the \textbf{$s \rightarrow h$} mode and the checkpoint trained by the authors on Parabank2.

\section{Details of retraining infilling models}
\label{sec:appendix_retrain}
We retrain the infilling model on summaries generated by MV-BART \citep{chen-yang-2020-multi}. The original approach uses the source document to train the infilling model and then makes predictions on the reference summary, which is to enhance the diversity of the pseudo data. However, we find that most of the subjects and objects extracted from the source dialogues are first- and second-person pronouns, such as "I" and "you", which are too different from the summaries from the third-person perspective. In order to adapt this approach to dialogue summarization, instead of using source documents, we use summaries generated by a model as training data for the infilling model.

\section{Model Settings and Training Details}
\label{sec:appendix_model_set}
Many FEC methods involve the construction of pseudo data. When it comes to data augmentation based on reference summaries and source documents, we use the training and validation sets from the summarization datasets SAMSum and DialogSum rather than our annotated data.

For different data augmentation approaches (\textbf{rule}, \textbf{infill}, and \textbf{infill-r}), we uniformly concatenate the summary to be corrected and the source document as input, and fine-tune some pre-trained models with the corrected summary as output for the above approaches. We conduct separate experiments using BART \footnote{using checkpoint from \url{https://huggingface.co/facebook/bart-large}}, PEGASUS \footnote{using checkpoint from \url{https://huggingface.co/sshleifer/distill-pegasus-cnn-16-4}} \citep{zhang2020pegasus} , T5  \footnote{using checkpoint from \url{https://huggingface.co/t5-base}} \citep{raffel2020exploring}. For all training modes, we fine-tune the pre-trained language models for 20 epochs with a batch size of 32, and use the loss on the validation set as the criterion for saving the best checkpoint. The learning rate is set to 3e-5. Hyperparameters for training the infilling models are kept at their default values.

When constructing pseudo data, \textbf{rule} generates 40451 and 35174 items on the training sets of SAMSum and DialogSum, and 2259 and 1369 items on the validation set of SAMSum and DialogSum. both \textbf{infill} and \textbf{infill-r} generate more pseudo data than \textbf{rule}. We randomly sample the pseudo data generated from \textbf{infill} and \textbf{infill-r} to ensure that the number of pseudo-data is the same as \textbf{rule}.

For CCGS, we re-train the classifier according to the original approach. To reflect its effectiveness more comprehensively, in addition to BART, BERT \citep{devlin-etal-2019-bert} and RoBERTa \citep{liu2019roberta} are also used as pre-trained models for the classifier. Hyperparameters are kept at their default values.

For FactPegasus, we use three Spacy models (Version 2.2.4) to pre-process the text separately: \texttt{en\_core\_web\_sm}, \texttt{en\_core\_web\_md}, \texttt{en\_core\_web\_lg}.

We use GeForce GTX 1080 Ti with 12GB memory for training and inference. Each single training session is less than 12 hours.

\section{Additional Figures and Tables}
\label{sec:appendix_fig_tab}

\begin{figure*}
\centering
\includegraphics[width=1.0\textwidth]{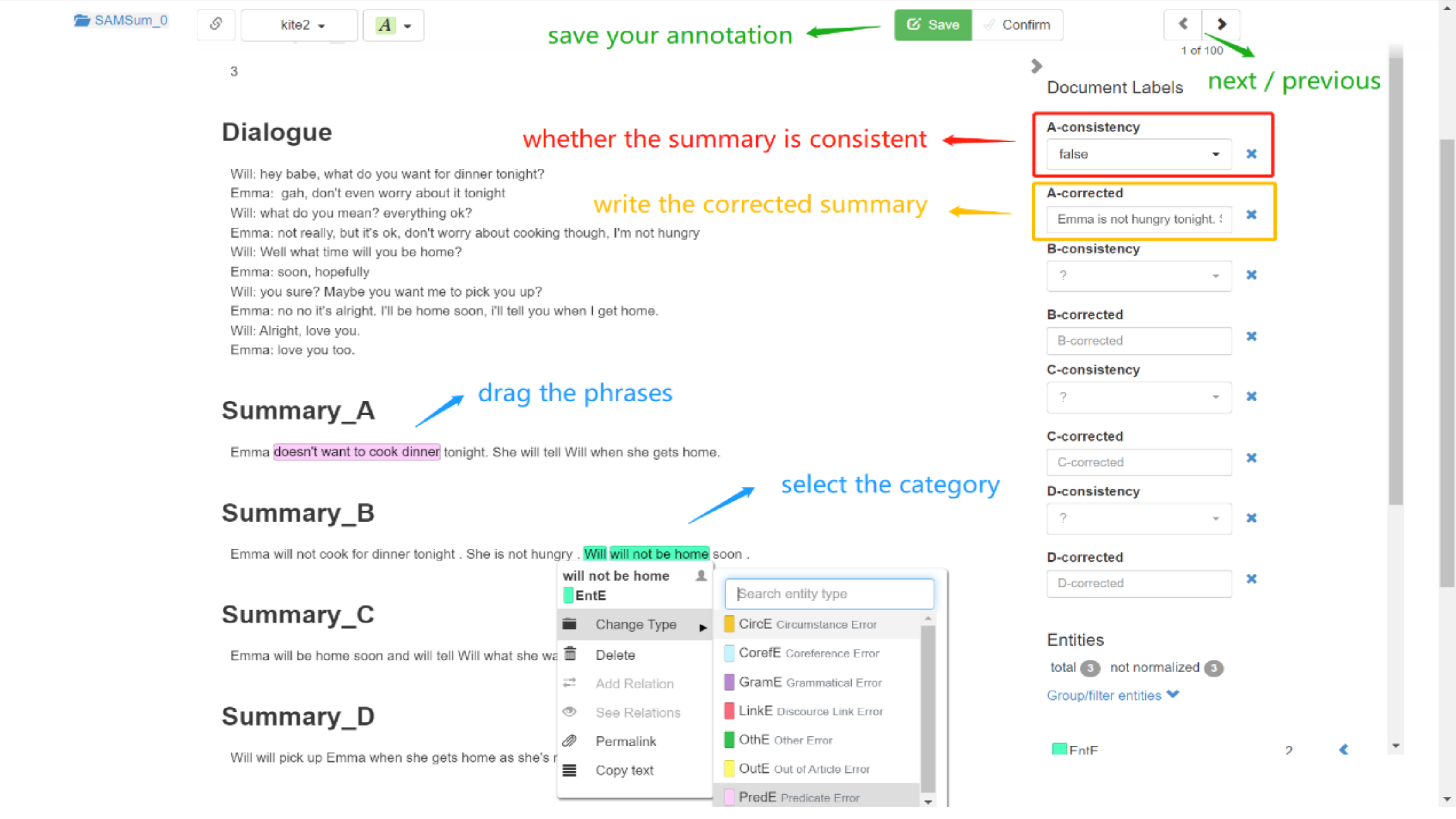}
\caption{Annotation Interface. }
\label{fig:interface}
\end{figure*}

\begin{figure*}
\centering
\includegraphics[width=1.0\textwidth]{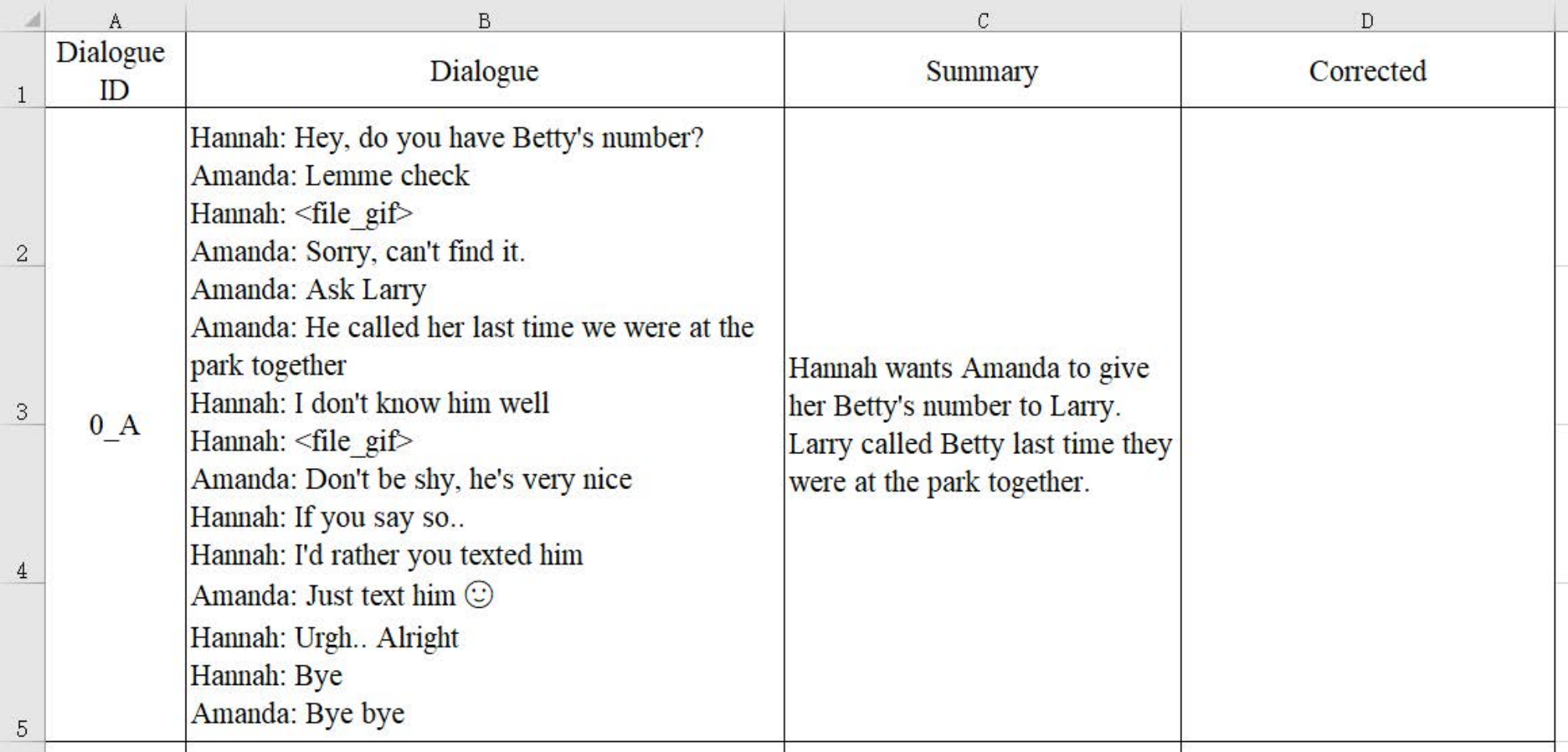}
\caption{The excel file for annotation. Annotators decide whether to use it according to their needs.}
\label{fig:interface_excel}
\end{figure*}


\begin{table*}[]
\resizebox{\linewidth}{!}{
\begin{tabular}{|crrrrr|}
\hline
\multicolumn{6}{|c|}{\textbf{SAMSum}} \\ \hline
\multicolumn{6}{|c|}{BART as the pre-trained   model} \\ \hline
\multicolumn{1}{|c|}{\textbf{Type}} & \multicolumn{1}{c}{\textbf{Pseudo}} & \multicolumn{1}{c}{\textbf{Real}} & \multicolumn{1}{c}{\textbf{Pseudo+Real}} & \multicolumn{1}{c}{\textbf{RefS}} & \multicolumn{1}{c|}{\textbf{Pseudo+RefS}} \\ \hline
\multicolumn{1}{|c|}{\textbf{FactCC}} & 0.2399 & 0.2365 & 0.2408 & 0.2220 & \textbf{0.2798} \\ \hline
\multicolumn{1}{|c|}{\textbf{DAE}} & 0.1776 & 0.1748 & \textbf{0.1783} & 0.1763 & 0.1687 \\ \hline
\multicolumn{1}{|c|}{\textbf{QuestEval}} & 0.4803 & 0.4760 & 0.4798 & \textbf{0.4863} & 0.4722 \\ \hline
\multicolumn{1}{|c|}{\textbf{BARTScore}} & -2.5505 & -2.6273 & -2.5510 & \textbf{-2.3863} & -2.4609 \\ \hline
\multicolumn{6}{|c|}{PEGASUS as the pre-trained   model} \\
\hline
\end{tabular}
\begin{tabular}{|crrrrr|}
\hline
\multicolumn{6}{|c|}{\textbf{DialogSum}} \\ \hline
\multicolumn{6}{|c|}{BART as the pre-trained   model} \\ \hline
\multicolumn{1}{|c|}{\textbf{Type}} & \multicolumn{1}{c}{\textbf{Pseudo}} & \multicolumn{1}{c}{\textbf{Real}} & \multicolumn{1}{c}{\textbf{Pseudo+Real}} & \multicolumn{1}{c}{\textbf{RefS}} & \multicolumn{1}{c|}{\textbf{Pseudo+RefS}} \\ \hline
\multicolumn{1}{|c|}{\textbf{FactCC}} & 0.1381 & 0.1298 & 0.1410 & \textbf{0.2396} & 0.2238 \\ \hline
\multicolumn{1}{|c|}{\textbf{DAE}} & 0.0754 & 0.0958 & 0.0883 & \textbf{0.1094} & 0.1050 \\ \hline
\multicolumn{1}{|c|}{\textbf{QuestEval}} & 0.3757 & \textbf{0.3775} & 0.3764 & 0.3687 & 0.3647 \\ \hline
\multicolumn{1}{|c|}{\textbf{BARTScore}} & -2.7102 & -2.7467 & -2.7283 & \textbf{-2.2739} & -2.4208 \\ \hline
\multicolumn{6}{|c|}{PEGASUS as the pre-trained   model} \\
\hline
\end{tabular}
}
\caption{
Performance (factuality metrics) of different training modes on SAMSum and DialogSum. The best results under the same pre-trained model are bolded. The data augmentation approach is set to \textbf{rule}. The pseudo data for the two datasets are constructed separately.
}
\label{tab:train_mode_samsum_dialogsum_fact}

\end{table*}

\begin{table*}[]
\resizebox{\linewidth}{!}{
\begin{tabular}{|crrrrrrrrrrrrrrr|}
\hline
\multicolumn{16}{|c|}{BART as the pre-trained model}                                                                                                                         \\ \hline
\multicolumn{1}{|c|}{} & \multicolumn{3}{c|}{\textbf{Pseudo}} & \multicolumn{3}{c|}{\textbf{Real}} & \multicolumn{3}{c|}{\textbf{Pseudo+Real}} & \multicolumn{3}{c|}{\textbf{RefS}} & \multicolumn{3}{c|}{\textbf{Pseudo+RefS}}                                                            \\ \hline
\multicolumn{1}{|c|}{\textbf{Type}} & \multicolumn{1}{c}{\textbf{P}} & \multicolumn{1}{c}{\textbf{R}} & \multicolumn{1}{c|}{\textbf{$\rm{F_{0.5}}$}} & \multicolumn{1}{c}{\textbf{P}} & \multicolumn{1}{c}{\textbf{R}} & \multicolumn{1}{c|}{\textbf{$\rm{F_{0.5}}$}} & \multicolumn{1}{c}{\textbf{P}} & \multicolumn{1}{c}{\textbf{R}} & \multicolumn{1}{c|}{\textbf{$\rm{F_{0.5}}$}} & \multicolumn{1}{c}{\textbf{P}} & \multicolumn{1}{c}{\textbf{R}} & \multicolumn{1}{c|}{\textbf{$\rm{F_{0.5}}$}} & \multicolumn{1}{c}{\textbf{P}} & \multicolumn{1}{c}{\textbf{R}} & \multicolumn{1}{c|}{\textbf{$\rm{F_{0.5}}$}} \\ \hline
\multicolumn{1}{|c|}{\textbf{M}} & 0.00 & 0.00 & \multicolumn{1}{r|}{0.00} & 0.00 & 0.00 & \multicolumn{1}{r|}{0.00} & 0.00 & 0.00 & \multicolumn{1}{r|}{0.00} & 1.74 & 9.30 & \multicolumn{1}{r|}{\textbf{2.08}} & 1.77 & 4.65 & 2.02                               \\
\multicolumn{1}{|c|}{\textbf{R}} & 6.38 & 1.83 & \multicolumn{1}{r|}{4.26} & 16.00 & 2.44 & \multicolumn{1}{r|}{7.58} & 26.47 & 5.49 & \multicolumn{1}{r|}{\textbf{15.00}} & 2.03 & 6.10 & \multicolumn{1}{r|}{2.34} & 1.25 & 3.66 & 1.44                               \\
\multicolumn{1}{|c|}{\textbf{U}} & 2.50 & 1.82 & \multicolumn{1}{r|}{7.04} & 6.25 & 5.45 & \multicolumn{1}{r|}{6.07} & 15.62 & 9.09 & \multicolumn{1}{r|}{\textbf{13.66}} & 3.40 & 9.09 & \multicolumn{1}{r|}{3.89} & 3.98 & 14.55 & 4.66                               \\
\hline
\end{tabular}
}
\caption{Performance (FERRANTI: form-based categories) of different training modes on SAMSum. The best $\rm{F_{0.5}}$ scores under the same pre-trained model are bolded. The data augmentation approach is set to \textbf{rule}.}
\label{tab:train_mode_samsum_form}
\end{table*}

\begin{table*}[]
\resizebox{\linewidth}{!}{
\begin{tabular}{|crrrrrrrrrrrrrrr|}
\hline
\multicolumn{16}{|c|}{BART as the pre-trained model}                                                                                                                                  \\ \hline
\multicolumn{1}{|c|}{} & \multicolumn{3}{c|}{\textbf{Pseudo}} & \multicolumn{3}{c|}{\textbf{Real}} & \multicolumn{3}{c|}{\textbf{Pseudo+Real}} & \multicolumn{3}{c|}{\textbf{RefS}} & \multicolumn{3}{c|}{\textbf{Pseudo+RefS}}                                                            \\ \hline
\multicolumn{1}{|c|}{\textbf{Type}} & \multicolumn{1}{c}{\textbf{TP}} & \multicolumn{1}{c}{\textbf{FP}} & \multicolumn{1}{c|}{\textbf{FN}} & \multicolumn{1}{c}{\textbf{TP}} & \multicolumn{1}{c}{\textbf{FP}} & \multicolumn{1}{c|}{\textbf{FN}} & \multicolumn{1}{c}{\textbf{TP}} & \multicolumn{1}{c}{\textbf{FP}} & \multicolumn{1}{c|}{\textbf{FN}} & \multicolumn{1}{c}{\textbf{TP}} & \multicolumn{1}{c}{\textbf{FP}} & \multicolumn{1}{c|}{\textbf{FN}} & \multicolumn{1}{c}{\textbf{TP}} & \multicolumn{1}{c}{\textbf{FP}} & \multicolumn{1}{c|}{\textbf{FN}} \\ \hline
\multicolumn{1}{|c|}{\textbf{M}} & 0 & 4 & \multicolumn{1}{r|}{43} & 0 & 17 & \multicolumn{1}{r|}{43} & 0 & 3 & \multicolumn{1}{r|}{43} & 4 & 226 & \multicolumn{1}{r|}{39} & 2 & 111 & 41                               \\
\multicolumn{1}{|c|}{\textbf{R}} & 3 & 44 & \multicolumn{1}{r|}{161} & 4 & 21 & \multicolumn{1}{r|}{160} & 9 & 25 & \multicolumn{1}{r|}{155} & 10 & 483 & \multicolumn{1}{r|}{154} & 6 & 474 & 158                              \\
\multicolumn{1}{|c|}{\textbf{U}} & 1 & 3 & \multicolumn{1}{r|}{54} & 3 & 45 & \multicolumn{1}{r|}{52} & 5 & 27 & \multicolumn{1}{r|}{50} & 5 & 142 & \multicolumn{1}{r|}{50} & 8 & 193 & 47                               \\
\multicolumn{1}{|c|}{\textbf{Total}} & 4 & 51 & \multicolumn{1}{r|}{258} & 7 & 83 & \multicolumn{1}{r|}{255} & 14 & 55 & \multicolumn{1}{r|}{248} & 19 & 851 & \multicolumn{1}{r|}{243} & 16 & 778 & 246                              \\
\hline
\end{tabular}
}
\caption{Performance (FERRANTI: form-based categories, TP, FP, FN) of different training modes on SAMSum. The data augmentation approach is set to \textbf{rule}.}
\label{tab:train_mode_samsum_tp_form}
\end{table*}

\begin{table*}[]
\resizebox{\linewidth}{!}{
\begin{tabular}{|crrrrrrrrrrrrrrr|}
\hline
\multicolumn{16}{|c|}{BART as the pre-trained model}                                                                                          \\ \hline
\multicolumn{1}{|c|}{} & \multicolumn{3}{c|}{\textbf{Pseudo}} & \multicolumn{3}{c|}{\textbf{Real}} & \multicolumn{3}{c|}{\textbf{Pseudo+Real}} & \multicolumn{3}{c|}{\textbf{RefS}} & \multicolumn{3}{c|}{\textbf{Pseudo+RefS}}                                                            \\ \hline
\multicolumn{1}{|c|}{\textbf{Type}} & \multicolumn{1}{c}{\textbf{P}} & \multicolumn{1}{c}{\textbf{R}} & \multicolumn{1}{c|}{\textbf{$\rm{F_{0.5}}$}} & \multicolumn{1}{c}{\textbf{P}} & \multicolumn{1}{c}{\textbf{R}} & \multicolumn{1}{c|}{\textbf{$\rm{F_{0.5}}$}} & \multicolumn{1}{c}{\textbf{P}} & \multicolumn{1}{c}{\textbf{R}} & \multicolumn{1}{c|}{\textbf{$\rm{F_{0.5}}$}} & \multicolumn{1}{c}{\textbf{P}} & \multicolumn{1}{c}{\textbf{R}} & \multicolumn{1}{c|}{\textbf{$\rm{F_{0.5}}$}} & \multicolumn{1}{c}{\textbf{P}} & \multicolumn{1}{c}{\textbf{R}} & \multicolumn{1}{c|}{\textbf{$\rm{F_{0.5}}$}} \\ \hline
\multicolumn{1}{|c|}{\textbf{M}} & 11.11 & 1.82 & \multicolumn{1}{r|}{\textbf{5.49}} & 0.00 & 0.00 & \multicolumn{1}{r|}{0.00} & 0.00 & 0.00 & \multicolumn{1}{r|}{0.00} & 0.66 & 1.82 & \multicolumn{1}{r|}{0.75} & 2.13 & 3.64 & 2.32                               \\
\multicolumn{1}{|c|}{\textbf{R}} & 4.32 & 1.96 & \multicolumn{1}{r|}{\textbf{3.48}} & 3.57 & 0.56 & \multicolumn{1}{r|}{1.72} & 3.70 & 0.56 & \multicolumn{1}{r|}{1.74} & 1.43 & 2.79 & \multicolumn{1}{r|}{1.58} & 1.20 & 2.51 & 1.34                               \\
\multicolumn{1}{|c|}{\textbf{U}} & 20.00 & 4.65 & \multicolumn{1}{r|}{\textbf{12.05}} & 3.95 & 6.98 & \multicolumn{1}{r|}{4.32} & 4.55 & 4.65 & \multicolumn{1}{r|}{4.57} & 2.52 & 13.95 & \multicolumn{1}{r|}{3.02} & 2.03 & 11.63 & 2.43                               \\
\hline
\end{tabular}
}
\caption{Performance (FERRANTI: form-based categories) of different training modes on DialogSum. The best $\rm{F_{0.5}}$ scores under the same pre-trained model are bolded. The data augmentation approach is set to \textbf{rule}.}
\label{tab:train_mode_dialogsum_form}
\end{table*}

\begin{table*}[]
\resizebox{\linewidth}{!}{
\begin{tabular}{|crrrrrrrrrrrrrrr|}
\hline
\multicolumn{16}{|c|}{BART as the pre-trained model}                                                                                 \\ \hline
\multicolumn{1}{|c|}{} & \multicolumn{3}{c|}{\textbf{Pseudo}} & \multicolumn{3}{c|}{\textbf{Real}} & \multicolumn{3}{c|}{\textbf{Pseudo+Real}} & \multicolumn{3}{c|}{\textbf{RefS}} & \multicolumn{3}{c|}{\textbf{Pseudo+RefS}}                                                            \\ \hline
\multicolumn{1}{|c|}{\textbf{Type}} & \multicolumn{1}{c}{\textbf{TP}} & \multicolumn{1}{c}{\textbf{FP}} & \multicolumn{1}{c|}{\textbf{FN}} & \multicolumn{1}{c}{\textbf{TP}} & \multicolumn{1}{c}{\textbf{FP}} & \multicolumn{1}{c|}{\textbf{FN}} & \multicolumn{1}{c}{\textbf{TP}} & \multicolumn{1}{c}{\textbf{FP}} & \multicolumn{1}{c|}{\textbf{FN}} & \multicolumn{1}{c}{\textbf{TP}} & \multicolumn{1}{c}{\textbf{FP}} & \multicolumn{1}{c|}{\textbf{FN}} & \multicolumn{1}{c}{\textbf{TP}} & \multicolumn{1}{c}{\textbf{FP}} & \multicolumn{1}{c|}{\textbf{FN}} \\ \hline
\multicolumn{1}{|c|}{\textbf{M}} & 1 & 8 & \multicolumn{1}{r|}{54} & 0 & 11 & \multicolumn{1}{r|}{55} & 0 & 4 & \multicolumn{1}{r|}{55} & 1 & 151 & \multicolumn{1}{r|}{54} & 2 & 92 & 53                               \\
\multicolumn{1}{|c|}{\textbf{R}} & 7 & 155 & \multicolumn{1}{r|}{351} & 2 & 54 & \multicolumn{1}{r|}{356} & 2 & 52 & \multicolumn{1}{r|}{356} & 10 & 691 & \multicolumn{1}{r|}{348} & 9 & 741 & 349                              \\
\multicolumn{1}{|c|}{\textbf{U}} & 2 & 8 & \multicolumn{1}{r|}{41} & 3 & 73 & \multicolumn{1}{r|}{40} & 2 & 42 & \multicolumn{1}{r|}{41} & 6 & 232 & \multicolumn{1}{r|}{37} & 5 & 241 & 38                               \\
\multicolumn{1}{|c|}{\textbf{Total}} & 10 & 171 & \multicolumn{1}{r|}{446} & 5 & 138 & \multicolumn{1}{r|}{451} & 4 & 98 & \multicolumn{1}{r|}{452} & 17 & 1074 & \multicolumn{1}{r|}{439} & 16 & 1074 & 440                              \\
\hline
\end{tabular}
}
\caption{Performance (FERRANTI: form-based categories, TP, FP, FN) of different training modes on DialogSum. The data augmentation approach is set to \textbf{rule}.}
\label{tab:train_mode_dialogsum_tp_form}
\end{table*}

\begin{table*}[]
\resizebox{\linewidth}{!}{
\begin{tabular}{|c|rrrrr|rrrrr|rrrrr|}
\hline
\multicolumn{1}{|l|}{} & \multicolumn{5}{c|}{BART as the pre-trained model} & \multicolumn{5}{c|}{PEGASUS as the pre-trained model} & \multicolumn{5}{c|}{T5 as the pre-trained model} \\ \cline{2-16} 
\multicolumn{1}{|l|}{} & \multicolumn{1}{c}{\textbf{Pseudo}} & \multicolumn{1}{c}{\textbf{Real}} & \multicolumn{1}{c}{\textbf{Pseudo+Real}} & \multicolumn{1}{c}{\textbf{RefS}} & \multicolumn{1}{c|}{\textbf{Pseudo+RefS}} & \multicolumn{1}{c}{\textbf{Pseudo}} & \multicolumn{1}{c}{\textbf{Real}} & \multicolumn{1}{c}{\textbf{Pseudo+Real}} & \multicolumn{1}{c}{\textbf{RefS}} & \multicolumn{1}{c|}{\textbf{Pseudo+RefS}} & \multicolumn{1}{c}{\textbf{Pseudo}} & \multicolumn{1}{c}{\textbf{Real}} & \multicolumn{1}{c}{\textbf{Pseudo+Real}} & \multicolumn{1}{c}{\textbf{RefS}} & \multicolumn{1}{c|}{\textbf{Pseudo+RefS}} \\ \hline
\textbf{Ent:ObjE} & 2.02 & 9.09 & \textbf{10.59} & 4.79 & 3.29 & 14.71 & 2.40 & \textbf{16.83} & 3.91 & 8.17 & 14.34 & 0.00 & \textbf{16.20} & 4.46 & 6.20 \\ \hline
\textbf{Ent:AttrE} & 0.00 & 0.00 & 0.00 & \textbf{5.81} & 0.00 & 0.00 & 0.00 & 0.00 & 0.00 & 0.00 & 0.00 & 0.00 & 0.00 & 0.00 & 0.00 \\ \hline
\textbf{Pred:ModE} & 0.00 & 0.00 & 0.00 & 0.00 & 0.00 & 0.00 & 0.00 & 0.00 & 0.00 & 0.00 & 0.00 & 0.00 & 0.00 & 0.00 & 0.00 \\ \hline
\textbf{Pred:TensE} & - & - & - & 0.00 & 0.00 & - & - & - & 0.00 & 0.00 & - & - & - & 0.00 & 0.00 \\ \hline
\textbf{Pred:NegE} & 0.00 & 0.00 & 0.00 & 0.00 & \textbf{21.74} & 0.00 & 0.00 & 0.00 & 0.00 & 0.00 & 0.00 & 0.00 & 0.00 & 0.00 & \textbf{33.33} \\ \hline
\textbf{Pred:VerbE} & 4.76 & 3.46 & \textbf{12.12} & 0.50 & 1.10 & 0.00 & \textbf{1.62} & 0.00 & 0.34 & 2.21 & 0.00 & \textbf{14.29} & 13.76 & 2.94 & 1.88 \\
\hline
\end{tabular}
}
\caption{Performance (FERRANTI: content-based categories) of different training modes on SAMSum. The values are all $\rm{F_{0.5}}$ scores. The best results under the same pre-trained model are bolded. The data augmentation approach is set to \textbf{rule}.}
\label{tab:train_mode_samsum_content}
\end{table*}

\begin{table*}[]
\resizebox{\linewidth}{!}{
\begin{tabular}{|c|rrrrr|rrrrr|rrrrr|}
\hline
\multicolumn{1}{|l|}{} & \multicolumn{5}{c|}{BART as the pre-trained model} & \multicolumn{5}{c|}{PEGASUS as the pre-trained model} & \multicolumn{5}{c|}{T5 as the pre-trained model} \\ \cline{2-16} 
\multicolumn{1}{|l|}{} & \multicolumn{1}{c}{\textbf{Pseudo}} & \multicolumn{1}{c}{\textbf{Real}} & \multicolumn{1}{c}{\textbf{Pseudo+Real}} & \multicolumn{1}{c}{\textbf{RefS}} & \multicolumn{1}{c|}{\textbf{Pseudo+RefS}} & \multicolumn{1}{c}{\textbf{Pseudo}} & \multicolumn{1}{c}{\textbf{Real}} & \multicolumn{1}{c}{\textbf{Pseudo+Real}} & \multicolumn{1}{c}{\textbf{RefS}} & \multicolumn{1}{c|}{\textbf{Pseudo+RefS}} & \multicolumn{1}{c}{\textbf{Pseudo}} & \multicolumn{1}{c}{\textbf{Real}} & \multicolumn{1}{c}{\textbf{Pseudo+Real}} & \multicolumn{1}{c}{\textbf{RefS}} & \multicolumn{1}{c|}{\textbf{Pseudo+RefS}} \\ \hline
\textbf{Ent:ObjE} & \textbf{5.17} & 2.54 & 3.90 & 3.89 & 4.68 & \textbf{15.59} & 10.99 & 8.16 & 5.60 & 6.05 & 11.79 & 4.88 & \textbf{16.33} & 6.51 & 8.32 \\ \hline
\textbf{Ent:AttrE} & 0.00 & 0.00 & 0.00 & 0.00 & 0.00 & 0.00 & 0.00 & 0.00 & 0.00 & 0.00 & 0.00 & 0.00 & 0.00 & \textbf{5.05} & 0.00 \\ \hline
\textbf{Pred:ModE} & 0.00 & 0.00 & 0.00 & 0.00 & 0.00 & 0.00 & 0.00 & 0.00 & \textbf{35.71} & 0.00 & 0.00 & 0.00 & 0.00 & \textbf{16.67} & 0.00 \\ \hline
\textbf{Pred:TensE} & - & - & - & 0.00 & - & - & 0.00 & - & 0.00 & - & - & - & - & 0.00 & - \\ \hline
\textbf{Pred:NegE} & - & - & - & 0.00 & - & - & - & - & - & - & - & - & - & 0.00 & - \\ \hline
\textbf{Pred:VerbE} & 0.00 & \textbf{2.77} & 0.00 & 1.12 & 0.74 & \textbf{2.07} & 1.20 & 0.00 & 0.06 & 0.14 & 0.00 & \textbf{2.15} & \textbf{2.15} & 1.09 & 0.80 \\
\hline
\end{tabular}
}
\caption{Performance (FERRANTI: content-based categories) of different training modes on DialogSum. The values are all $\rm{F_{0.5}}$ scores. The best results under the same pre-trained model are bolded. The data augmentation approach is set to \textbf{rule}.}
\label{tab:train_mode_dialogsum_content}
\end{table*}


\begin{table*}[]
\resizebox{\linewidth}{!}{
\begin{tabular}{|c|rrrrrr|rrrrrr|rrrrrr|}
\hline
 & \multicolumn{6}{c|}{BART as pre-trained models} & \multicolumn{6}{c|}{PEGASUS as pre-trained models} & \multicolumn{6}{c|}{T5 as pre-trained models} \\ \cline{2-19} 
 & \multicolumn{3}{c}{\textbf{Pseudo}} & \multicolumn{3}{c|}{\textbf{Pseudo+Real}} & \multicolumn{3}{c}{\textbf{Pseudo}} & \multicolumn{3}{c|}{\textbf{Pseudo+Real}} & \multicolumn{3}{c}{\textbf{Pseudo}} & \multicolumn{3}{c|}{\textbf{Pseudo+Real}} \\ \cline{2-19} 
 & \multicolumn{1}{c}{\textbf{rule}} & \multicolumn{1}{c}{\textbf{infill}} & \multicolumn{1}{c}{\textbf{infill-r}} & \multicolumn{1}{c}{\textbf{rule}} & \multicolumn{1}{c}{\textbf{infill}} & \multicolumn{1}{c|}{\textbf{infill-r}} & \multicolumn{1}{c}{\textbf{rule}} & \multicolumn{1}{c}{\textbf{Infill}} & \multicolumn{1}{c}{\textbf{infill-r}} & \multicolumn{1}{c}{\textbf{rule}} & \multicolumn{1}{c}{\textbf{Infill}} & \multicolumn{1}{c|}{\textbf{infill-r}} & \multicolumn{1}{c}{\textbf{rule}} & \multicolumn{1}{c}{\textbf{infill}} & \multicolumn{1}{c}{\textbf{infill-r}} & \multicolumn{1}{c}{\textbf{rule}} & \multicolumn{1}{c}{\textbf{infill}} & \multicolumn{1}{c|}{\textbf{infill-r}} \\ \hline
\textbf{FactCC} & 0.2399 & 0.2370 & 0.2426 & 0.2408 & \textbf{0.2432} & 0.2370 & 0.2349 & 0.2362 & 0.2344 & 0.2373 & 0.2323 & \textbf{0.2386} & 0.2380 & 0.2374 & \textbf{0.2415} & 0.2395 & 0.2411 & 0.2395 \\
\textbf{DAE} & 0.1776 & 0.1754 & 0.1757 & \textbf{0.1783} & 0.1779 & 0.1737 & 0.1837 & 0.1766 & 0.1805 & 0.1796 & \textbf{0.1866} & 0.1809 & 0.1800 & 0.1800 & 0.1768 & 0.1812 & \textbf{0.1829} & 0.1812 \\
\textbf{QuestEval} & 0.4803 & \textbf{0.4808} & 0.4774 & 0.4798 & 0.4785 & 0.4778 & \textbf{0.4794} & 0.4793 & 0.4793 & 0.4790 & 0.4789 & 0.4790 & 0.4819 & 0.4784 & 0.4797 & \textbf{0.4824} & 0.4808 & 0.4796 \\
\hline
\end{tabular}
}
\caption{Performance (factuality metrics) of different data augmentation approaches on SAMSum. The best results under the same pre-trained model are bolded.}
\label{tab:approach_samsum_fact}
\end{table*}

\begin{table*}[]
\resizebox{\linewidth}{!}{
\begin{tabular}{|c|rrrrrr|rrrrrr|rrrrrr|}
\hline
 & \multicolumn{6}{c|}{BART as pre-trained models} & \multicolumn{6}{c|}{PEGASUS as pre-trained models} & \multicolumn{6}{c|}{T5 as pre-trained models} \\ \cline{2-19} 
 & \multicolumn{3}{c}{\textbf{Pseudo}} & \multicolumn{3}{c|}{\textbf{Pseudo+Real}} & \multicolumn{3}{c}{\textbf{Pseudo}} & \multicolumn{3}{c|}{\textbf{Pseudo+Real}} & \multicolumn{3}{c}{\textbf{Pseudo}} & \multicolumn{3}{c|}{\textbf{Pseudo+Real}} \\ \cline{2-19} 
 & \multicolumn{1}{c}{\textbf{rule}} & \multicolumn{1}{c}{\textbf{infill}} & \multicolumn{1}{c}{\textbf{infill-r}} & \multicolumn{1}{c}{\textbf{rule}} & \multicolumn{1}{c}{\textbf{infill}} & \multicolumn{1}{c|}{\textbf{infill-r}} & \multicolumn{1}{c}{\textbf{rule}} & \multicolumn{1}{c}{\textbf{Infill}} & \multicolumn{1}{c}{\textbf{infill-r}} & \multicolumn{1}{c}{\textbf{rule}} & \multicolumn{1}{c}{\textbf{Infill}} & \multicolumn{1}{c|}{\textbf{infill-r}} & \multicolumn{1}{c}{\textbf{rule}} & \multicolumn{1}{c}{\textbf{infill}} & \multicolumn{1}{c}{\textbf{infill-r}} & \multicolumn{1}{c}{\textbf{rule}} & \multicolumn{1}{c}{\textbf{infill}} & \multicolumn{1}{c|}{\textbf{infill-r}} \\ \hline
\textbf{FactCC} & 0.1381 & 0.1397 & 0.1347 & \textbf{0.1410} & 0.1298 & 0.1327 & 0.1366 & 0.1297 & 0.1305 & 0.1348 & \textbf{0.1379} & 0.1364 & \textbf{0.1479} & 0.1309 & 0.1309 & 0.1322 & 0.1310 & 0.1297 \\
\textbf{DAE} & 0.0754 & 0.0838 & 0.0858 & 0.0883 & 0.0954 & \textbf{0.0996} & 0.0890 & 0.0879 & 0.0883 & \textbf{0.0967} & 0.0958 & 0.0946 & 0.0877 & 0.0921 & 0.0921 & \textbf{0.0933} & 0.0896 & 0.0921 \\
\textbf{QuestEval} & 0.3757 & \textbf{0.3774} & 0.3765 & 0.3764 & 0.3765 & 0.3771 & 0.3758 & \textbf{0.3795} & 0.3783 & 0.3782 & 0.3788 & 0.3786 & 0.3759 & 0.3780 & 0.3774 & 0.3780 & 0.3777 & \textbf{0.3783} \\
\hline
\end{tabular}
}
\caption{Performance (factuality metrics) of different data augmentation approaches on DialogSum. The best results under the same pre-trained model are bolded.}
\label{tab:approach_dialogsum_fact}
\end{table*}

\begin{table*}[]
\resizebox{\linewidth}{!}{
\begin{tabular}{|cc|rrrrrr|rrrrrr|rrrrrr|}
\hline
\multicolumn{2}{|l|}{\multirow{3}{*}{}} & \multicolumn{6}{c|}{BART as the pre-trained model} & \multicolumn{6}{c|}{PEGASUS as the pre-trained model} & \multicolumn{6}{c|}{T5 as the pre-trained model}                                                                                                                                                                                      \\ \cline{3-20} 
\multicolumn{2}{|l|}{} & \multicolumn{3}{c}{\textbf{Pseudo}} & \multicolumn{3}{c|}{\textbf{Pseudo+Real}} & \multicolumn{3}{c}{\textbf{Pseudo}} & \multicolumn{3}{c|}{\textbf{Pseudo+Real}} & \multicolumn{3}{c}{\textbf{Pseudo}} & \multicolumn{3}{c|}{\textbf{Pseudo+Real}}                                                                        \\ \cline{3-20} 
\multicolumn{2}{|l|}{} & \multicolumn{1}{c}{\textbf{rule}} & \multicolumn{1}{c}{\textbf{infill}} & \multicolumn{1}{c}{\textbf{infill-r}} & \multicolumn{1}{c}{\textbf{rule}} & \multicolumn{1}{c}{\textbf{infill}} & \multicolumn{1}{c|}{\textbf{infill-r}} & \multicolumn{1}{c}{\textbf{rule}} & \multicolumn{1}{c}{\textbf{Infill}} & \multicolumn{1}{c}{\textbf{infill-r}} & \multicolumn{1}{c}{\textbf{rule}} & \multicolumn{1}{c}{\textbf{Infill}} & \multicolumn{1}{c|}{\textbf{infill-r}} & \multicolumn{1}{c}{\textbf{rule}} & \multicolumn{1}{c}{\textbf{infill}} & \multicolumn{1}{c}{\textbf{infill-r}} & \multicolumn{1}{c}{\textbf{rule}} & \multicolumn{1}{c}{\textbf{infill}} & \multicolumn{1}{c|}{\textbf{infill-r}} \\ \hline
\multicolumn{1}{|c|}{\multirow{3}{*}{\textbf{Ent:ObjE}}} & \textbf{P} & 2.78 & 22.22 & 7.41 & 15.15 & 13.79 & 15.38 & 24.00 & 9.09 & 18.75 & 26.92 & 25.00 & 23.08 & 20.00 & 20.00 & 10.00 & 25.00 & 10.00 & 14.29                                  \\
\multicolumn{1}{|c|}{} & \textbf{R} & 0.96 & 3.85 & 1.92 & 4.81 & 3.85 & 3.85 & 5.77 & 1.92 & 2.88 & 6.73 & 3.85 & 2.88 & 6.73 & 0.96 & 0.96 & 6.73 & 0.96 & 2.56                                   \\
\hline
\end{tabular}
}
\caption{Performance (FERRANTI: content-based categories) of different data augmentation approaches on SAMSum. The best $\rm{F_{0.5}}$ scores under the same pre-trained model are bolded.}
\label{tab:approach_samsum_content}
\end{table*}

\begin{table*}[]
\resizebox{\linewidth}{!}{
\begin{tabular}{|cc|rrrrrr|rrrrrr|rrrrrr|}
\hline
\multicolumn{2}{|l|}{\multirow{3}{*}{}} & \multicolumn{6}{c|}{BART as the pre-trained model} & \multicolumn{6}{c|}{PEGASUS as the pre-trained model} & \multicolumn{6}{c|}{T5 as the pre-trained model}                                                                                                        \\ \cline{3-20} 
\multicolumn{2}{|l|}{} & \multicolumn{3}{c}{\textbf{Pseudo}} & \multicolumn{3}{c|}{\textbf{Pseudo+Real}} & \multicolumn{3}{c}{\textbf{Pseudo}} & \multicolumn{3}{c|}{\textbf{Pseudo+Real}} & \multicolumn{3}{c}{\textbf{Pseudo}} & \multicolumn{3}{c|}{\textbf{Pseudo+Real}}                                                                        \\ \cline{3-20} 
\multicolumn{2}{|l|}{} & \multicolumn{1}{c}{\textbf{rule}} & \multicolumn{1}{c}{\textbf{infill}} & \multicolumn{1}{c}{\textbf{infill-r}} & \multicolumn{1}{c}{\textbf{rule}} & \multicolumn{1}{c}{\textbf{infill}} & \multicolumn{1}{c|}{\textbf{infill-r}} & \multicolumn{1}{c}{\textbf{rule}} & \multicolumn{1}{c}{\textbf{infill}} & \multicolumn{1}{c}{\textbf{infill-r}} & \multicolumn{1}{c}{\textbf{rule}} & \multicolumn{1}{c}{\textbf{infill}} & \multicolumn{1}{c|}{\textbf{infill-r}} & \multicolumn{1}{c}{\textbf{rule}} & \multicolumn{1}{c}{\textbf{infill}} & \multicolumn{1}{c}{\textbf{infill-r}} & \multicolumn{1}{c}{\textbf{rule}} & \multicolumn{1}{c}{\textbf{infill}} & \multicolumn{1}{c|}{\textbf{infill-r}} \\ \hline
\multicolumn{1}{|c|}{\multirow{3}{*}{\textbf{Ent:ObjE}}} & \textbf{TP} & 1 & 4 & 2 & 5 & 4 & 4 & 6 & 2 & 3 & 7 & 4 & 3 & 7 & 1 & 1 & 7 & 1 & 1                                      \\
\multicolumn{1}{|c|}{} & \textbf{FP} & 35 & 14 & 25 & 28 & 25 & 22 & 19 & 20 & 13 & 19 & 12 & 10 & 28 & 4 & 9 & 21 & 9 & 9                                      \\
\multicolumn{1}{|c|}{} & \textbf{FN} & 103 & 100 & 102 & 99 & 100 & 100 & 98 & 102 & 101 & 97 & 100 & 101 & 97 & 103 & 103 & 97 & 103 & 103                                    \\
\hline
\end{tabular}
}
\caption{Performance (FERRANTI: content-based categories, TP, FP, FN) of different data augmentation approaches on SAMSum.}
\label{tab:approach_samsum_tp_content}
\end{table*}

\begin{table*}[]
\resizebox{\linewidth}{!}{
\begin{tabular}{|c|rrrrrr|rrrrrr|rrrrrr|}
\hline
\multicolumn{1}{|l|}{\multirow{3}{*}{}} & \multicolumn{6}{c|}{BART as the pre-trained model} & \multicolumn{6}{c|}{PEGASUS as the pre-trained model} & \multicolumn{6}{c|}{T5 as the pre-trained model} \\ \cline{2-19} 
\multicolumn{1}{|l|}{} & \multicolumn{3}{c}{\textbf{Pseudo}} & \multicolumn{3}{c|}{\textbf{Pseudo+Real}} & \multicolumn{3}{c}{\textbf{Pseudo}} & \multicolumn{3}{c|}{\textbf{Pseudo+Real}} & \multicolumn{3}{c}{\textbf{Pseudo}} & \multicolumn{3}{c|}{\textbf{Pseudo+Real}} \\ \cline{2-19} 
\multicolumn{1}{|l|}{} & \multicolumn{1}{c}{\textbf{rule}} & \multicolumn{1}{c}{\textbf{infill}} & \multicolumn{1}{c}{\textbf{infill-r}} & \multicolumn{1}{c}{\textbf{rule}} & \multicolumn{1}{c}{\textbf{infill}} & \multicolumn{1}{c|}{\textbf{infill-r}} & \multicolumn{1}{c}{\textbf{rule}} & \multicolumn{1}{c}{\textbf{infill}} & \multicolumn{1}{c}{\textbf{infill-r}} & \multicolumn{1}{c}{\textbf{rule}} & \multicolumn{1}{c}{\textbf{infill}} & \multicolumn{1}{c|}{\textbf{infill-r}} & \multicolumn{1}{c}{\textbf{rule}} & \multicolumn{1}{c}{\textbf{infill}} & \multicolumn{1}{c}{\textbf{infill-r}} & \multicolumn{1}{c}{\textbf{rule}} & \multicolumn{1}{c}{\textbf{infill}} & \multicolumn{1}{c|}{\textbf{infill-r}} \\ \hline
\textbf{Ent:ObjE} & 5.17 & \textbf{5.82} & 4.46 & 3.90 & 4.23 & 5.07 & \textbf{15.59} & 4.22 & 4.37 & 8.16 & 14.87 & 11.67 & 11.79 & 2.49 & 2.39 & \textbf{16.33} & 13.41 & 11.67 \\ \hline
\textbf{Ent:AttrE} & 0.00 & 0.00 & 0.00 & 0.00 & 0.00 & 0.00 & 0.00 & 0.00 & 0.00 & 0.00 & 0.00 & 0.00 & 0.00 & 0.00 & 0.00 & 0.00 & 0.00 & 0.00 \\ \hline
\textbf{Pred:ModE} & 0.00 & 0.00 & 0.00 & 0.00 & 0.00 & 0.00 & 0.00 & 0.00 & 0.00 & 0.00 & 0.00 & 0.00 & 0.00 & 0.00 & 0.00 & 0.00 & 0.00 & 0.00 \\ \hline
\textbf{Pred:TensE} & - & - & - & - & - & - & - & - & - & - & - & - & - & - & 0.00 & - & - & - \\ \hline
\textbf{Pred:NegE} & - & - & - & - & - & - & - & - & - & - & - & - & - & - & - & - & - & - \\
\hline
\end{tabular}
}
\caption{Performance (FERRANTI: content-based categories) of different data augmentation approaches on DialogSum. The values are all $\rm{F_{0.5}}$ scores. The best results under the same pre-trained model are bolded. The pseudo-data corpus is DialogSum.  The pseudo-data corpus is DialogSum. Please see Table \ref{tab:approach_dialogsum_content} and Table \ref{tab:approach_dialogsum_tp_content} in Appendix \ref{sec:appendix_fig_tab} for precision, recall, or TP, etc.}
\label{tab:approach_dialogsum_short_content}
\end{table*}

\begin{table*}[]
\resizebox{\linewidth}{!}{
\begin{tabular}{|cc|llllll|llllll|llllll|}
\hline
\multicolumn{2}{|l|}{\multirow{3}{*}{}} & \multicolumn{6}{c|}{BART as the pre-trained model} & \multicolumn{6}{c|}{PEGASUS as the pre-trained model} & \multicolumn{6}{c|}{T5 as the pre-trained model} \\ \cline{3-20} 
\multicolumn{2}{|l|}{} & \multicolumn{3}{c}{\textbf{Pseudo}} & \multicolumn{3}{c|}{\textbf{Pseudo+Real}} & \multicolumn{3}{c}{\textbf{Pseudo}} & \multicolumn{3}{c|}{\textbf{Pseudo+Real}} & \multicolumn{3}{c}{\textbf{Pseudo}} & \multicolumn{3}{c|}{\textbf{Pseudo+Real}} \\ \cline{3-20} 
\multicolumn{2}{|l|}{} & \multicolumn{1}{c}{\textbf{rule}} & \multicolumn{1}{c}{\textbf{infill}} & \multicolumn{1}{c}{\textbf{infill-r}} & \multicolumn{1}{c}{\textbf{rule}} & \multicolumn{1}{c}{\textbf{infill}} & \multicolumn{1}{c|}{\textbf{infill-r}} & \multicolumn{1}{c}{\textbf{rule}} & \multicolumn{1}{c}{\textbf{Infill}} & \multicolumn{1}{c}{\textbf{infill-r}} & \multicolumn{1}{c}{\textbf{rule}} & \multicolumn{1}{c}{\textbf{Infill}} & \multicolumn{1}{c|}{\textbf{infill-r}} & \multicolumn{1}{c}{\textbf{rule}} & \multicolumn{1}{c}{\textbf{infill}} & \multicolumn{1}{c}{\textbf{infill-r}} & \multicolumn{1}{c}{\textbf{rule}} & \multicolumn{1}{c}{\textbf{infill}} & \multicolumn{1}{c|}{\textbf{infill-r}} \\ \hline
\multicolumn{1}{|c|}{\multirow{3}{*}{\textbf{Ent:ObjE}}} & \textbf{P} & 5.30 & 6.48 & 5.10 & 5.56 & 5.26 & 6.17 & 20.97 & 11.76 & 13.33 & 21.05 & 32.00 & 27.27 & 14.12 & 12.50 & 10.00 & 42.11 & 30.43 & 27.27 \\
\multicolumn{1}{|c|}{} & \textbf{R} & 4.73 & 4.14 & 2.96 & 1.78 & 2.37 & 2.96 & 7.69 & 1.18 & 1.18 & 2.37 & 4.73 & 3.55 & 7.10 & 0.59 & 0.59 & 4.73 & 4.14 & 3.55 \\
\multicolumn{1}{|c|}{} & \textbf{$\rm{F_{0.5}}$} & 5.17 & \textbf{5.82} & 4.46 & 3.90 & 4.23 & 5.07 & \textbf{15.59} & 4.22 & 4.37 & 8.16 & 14.87 & 11.67 & 11.79 & 2.49 & 2.39 & \textbf{16.33} & 13.41 & 11.67 \\
\hline
\end{tabular}
}
\caption{Performance (FERRANTI: content-based categories) of different data augmentation approaches on DialogSum. The best $\rm{F_{0.5}}$ scores under the same pre-trained model are bolded.}
\label{tab:approach_dialogsum_content}
\end{table*}

\begin{table*}[]
\resizebox{\linewidth}{!}{
\begin{tabular}{|cc|rrrrrr|rrrrrr|rrrrrr|}
\hline
\multicolumn{2}{|l|}{\multirow{3}{*}{}} & \multicolumn{6}{c|}{BART as the pre-trained model} & \multicolumn{6}{c|}{PEGASUS as the pre-trained model} & \multicolumn{6}{c|}{T5 as the pre-trained model} \\ \cline{3-20} 
\multicolumn{2}{|l|}{} & \multicolumn{3}{c}{\textbf{Pseudo}} & \multicolumn{3}{c|}{\textbf{Pseudo+Real}} & \multicolumn{3}{c}{\textbf{Pseudo}} & \multicolumn{3}{c|}{\textbf{Pseudo+Real}} & \multicolumn{3}{c}{\textbf{Pseudo}} & \multicolumn{3}{c|}{\textbf{Pseudo+Real}} \\ \cline{3-20} 
\multicolumn{2}{|l|}{} & \multicolumn{1}{c}{\textbf{rule}} & \multicolumn{1}{c}{\textbf{infill}} & \multicolumn{1}{c}{\textbf{infill-r}} & \multicolumn{1}{c}{\textbf{rule}} & \multicolumn{1}{c}{\textbf{infill}} & \multicolumn{1}{c|}{\textbf{infill-r}} & \multicolumn{1}{c}{\textbf{rule}} & \multicolumn{1}{c}{\textbf{infill}} & \multicolumn{1}{c}{\textbf{infill-r}} & \multicolumn{1}{c}{\textbf{rule}} & \multicolumn{1}{c}{\textbf{infill}} & \multicolumn{1}{c|}{\textbf{infill-r}} & \multicolumn{1}{c}{\textbf{rule}} & \multicolumn{1}{c}{\textbf{infill}} & \multicolumn{1}{c}{\textbf{infill-r}} & \multicolumn{1}{c}{\textbf{rule}} & \multicolumn{1}{c}{\textbf{infill}} & \multicolumn{1}{c|}{\textbf{infill-r}} \\ \hline
\multicolumn{1}{|c|}{\multirow{3}{*}{\textbf{Ent:ObjE}}} & \textbf{TP} & 8 & 7 & 5 & 3 & 4 & 5 & 13 & 2 & 2 & 4 & 8 & 6 & 12 & 1 & 1 & 8 & 7 & 6 \\
\multicolumn{1}{|c|}{} & \textbf{FP} & 143 & 101 & 93 & 51 & 72 & 76 & 49 & 15 & 13 & 15 & 17 & 16 & 73 & 7 & 9 & 11 & 16 & 16 \\
\multicolumn{1}{|c|}{} & \textbf{FN} & 161 & 162 & 164 & 166 & 165 & 164 & 156 & 167 & 167 & 165 & 161 & 163 & 157 & 168 & 168 & 161 & 162 & 163 \\ \hline
\multicolumn{1}{|c|}{\multirow{3}{*}{\textbf{Ent:AttrE}}} & \textbf{TP} & 0 & 0 & 0 & 0 & 0 & 0 & 0 & 0 & 0 & 0 & 0 & 0 & 0 & 0 & 0 & 0 & 0 & 0 \\
\multicolumn{1}{|c|}{} & \textbf{FP} & 1 & 2 & 2 & 5 & 2 & 2 & 0 & 5 & 4 & 0 & 0 & 0 & 0 & 0 & 0 & 0 & 2 & 0 \\
\hline
\end{tabular}
}
\caption{Performance (FERRANTI: content-based categories, TP, FP, FN) of different data augmentation approaches on DialogSum. }
\label{tab:approach_dialogsum_tp_content}
\end{table*}

\begin{table*}[]
\resizebox{\linewidth}{!}{
\begin{tabular}{|crrrrrr|}
\hline
\multicolumn{7}{|c|}{\textbf{SAMSum}} \\ \hline
\multicolumn{7}{|c|}{BART as the   pre-trained model} \\ \hline
\multicolumn{1}{|l|}{\multirow{2}{*}{}} & \multicolumn{3}{c}{\textbf{Pseudo}} & \multicolumn{3}{c|}{\textbf{Pseudo+Real}} \\ \cline{2-7} 
\multicolumn{1}{|l|}{} & \multicolumn{1}{c}{\textbf{rule}} & \multicolumn{1}{c}{\textbf{infill}} & \multicolumn{1}{c}{\textbf{inill-r}} & \multicolumn{1}{c}{\textbf{rule}} & \multicolumn{1}{c}{\textbf{infill}} & \multicolumn{1}{c|}{\textbf{inill-r}} \\ \hline
\multicolumn{1}{|c|}{\textbf{M}} & 0.00 & 0.00 & 0.00 & 0.00 & 0.00 & 0.00 \\
\multicolumn{1}{|c|}{\textbf{R}} & 4.26 & 4.63 & 1.36 & \textbf{15.00} & 9.62 & 8.12 \\
\multicolumn{1}{|c|}{\textbf{U}} & 7.04 & 11.49 & 13.33 & \textbf{13.66} & 7.41 & 13.25 \\
\multicolumn{1}{|c|}{\textbf{Total}} & 4.15 & 5.23 & 2.92 & \textbf{13.01} & 8.10 & 8.96 \\
\hline
\end{tabular}

\begin{tabular}{|crrrrrr|}
\hline
\multicolumn{7}{|c|}{\textbf{DialogSum}} \\ \hline
\multicolumn{7}{|c|}{BART as the pre-trained   model} \\ \hline
\multicolumn{1}{|c|}{\multirow{2}{*}{}} & \multicolumn{3}{c}{\textbf{Pseudo}} & \multicolumn{3}{c|}{\textbf{Pseudo+Real}} \\ \cline{2-7} 
\multicolumn{1}{|c|}{} & \multicolumn{1}{c}{\textbf{rule}} & \multicolumn{1}{c}{\textbf{infill}} & \multicolumn{1}{c}{\textbf{inill-r}} & \multicolumn{1}{c}{\textbf{rule}} & \multicolumn{1}{c}{\textbf{infill}} & \multicolumn{1}{c|}{\textbf{inill-r}} \\ \hline
\multicolumn{1}{|c|}{\textbf{M}} & 5.49 & 0.00 & \textbf{10.10} & 0.00 & 0.00 & 5.26 \\
\multicolumn{1}{|c|}{\textbf{R}} & 3.48 & \textbf{3.60} & 2.36 & 1.74 & 2.15 & 2.11 \\
\multicolumn{1}{|c|}{\textbf{U}} & \textbf{12.05} & 9.68 & 5.99 & 4.57 & 5.85 & 5.13 \\
\multicolumn{1}{|c|}{\textbf{Total}} & 4.24 & \textbf{4.28} & 3.60 & 2.31 & 2.63 & 3.00 \\
\hline
\end{tabular}
}

\caption{Performance (FERRANTI: form-based categories) of different data augmentation approaches on SAMSum and DialogSum. The values are all $\rm{F_{0.5}}$ scores. The best results under the same pre-trained model are bolded. The pseudo data for the two datasets are constructed separately.}
\label{tab:approach_samsum_dialogsum_short_form}
\end{table*}


\begin{table*}[]
\resizebox{\linewidth}{!}{
\centering
\begin{tabular}{|ccrrr|ccrrr|}
\hline
\multicolumn{5}{|c|}{\textbf{SAMSum}} & \multicolumn{5}{c|}{\textbf{DialogSum}} \\ \hline
\multicolumn{2}{|c|}{Pre-trained Models} & \multicolumn{1}{c}{BART} & \multicolumn{1}{c}{BERT} & \multicolumn{1}{c|}{RoBERTa} & \multicolumn{2}{c|}{Pre-trained Models} & \multicolumn{1}{c}{BART} & \multicolumn{1}{c}{BERT} & \multicolumn{1}{c|}{RoBERTa} \\ \hline
\multicolumn{1}{|c|}{\multirow{3}{*}{\textbf{Ent:ObjE}}} & \multicolumn{1}{c|}{\textbf{TP}} & 1 & 0 & 1 & \multicolumn{1}{c|}{\multirow{3}{*}{\textbf{Ent:ObjE}}} & \multicolumn{1}{c|}{\textbf{TP}} & 1 & 2 & 2 \\
\multicolumn{1}{|c|}{} & \multicolumn{1}{c|}{\textbf{FP}} & 20 & 28 & 36 & \multicolumn{1}{c|}{} & \multicolumn{1}{c|}{\textbf{FP}} & 25 & 20 & 12 \\
\multicolumn{1}{|c|}{} & \multicolumn{1}{c|}{\textbf{FN}} & 103 & 104 & 103 & \multicolumn{1}{c|}{} & \multicolumn{1}{c|}{\textbf{FN}} & 168 & 167 & 167 \\ \hline
\multicolumn{1}{|c|}{\multirow{3}{*}{\textbf{Ent:AttrE}}} & \multicolumn{1}{c|}{\textbf{TP}} & 0 & 0 & 0 & \multicolumn{1}{c|}{\multirow{3}{*}{\textbf{Ent:AttrE}}} & \multicolumn{1}{c|}{\textbf{TP}} & 0 & 0 & 0 \\
\multicolumn{1}{|c|}{} & \multicolumn{1}{c|}{\textbf{FP}} & 0 & 2 & 0 & \multicolumn{1}{c|}{} & \multicolumn{1}{c|}{\textbf{FP}} & 1 & 2 & 1 \\
\multicolumn{1}{|c|}{} & \multicolumn{1}{c|}{\textbf{FN}} & 6 & 6 & 6 & \multicolumn{1}{c|}{} & \multicolumn{1}{c|}{\textbf{FN}} & 7 & 7 & 7 \\
\hline
\end{tabular}
}
\caption{Performance (FERRANTI: content-based categories, TP, FP, FN) of CCGS on SAMSum and DialogSum.}
\label{tab:ccgs_samsum_dialogsum_tp_content}
\end{table*}

\begin{table*}[]
\resizebox{\linewidth}{!}{
\centering
\begin{tabular}{|ccrrr|ccrrr|}
\hline
\multicolumn{5}{|c|}{\textbf{SAMSum}} & \multicolumn{5}{c|}{\textbf{DialogSum}} \\ \hline
\multicolumn{2}{|c|}{Spacy Models} & \multicolumn{1}{c}{sm} & \multicolumn{1}{c}{md} & \multicolumn{1}{c|}{lg} & \multicolumn{2}{c|}{Spacy Models} & \multicolumn{1}{c}{sm} & \multicolumn{1}{c}{md} & \multicolumn{1}{c|}{lg} \\ \hline
\multicolumn{1}{|c|}{\multirow{3}{*}{\textbf{Ent:ObjE}}} & \multicolumn{1}{c|}{\textbf{TP}} & 3 & 2 & 5 & \multicolumn{1}{c|}{\multirow{3}{*}{\textbf{Ent:ObjE}}} & \multicolumn{1}{c|}{\textbf{TP}} & 1 & 2 & 1 \\
\multicolumn{1}{|c|}{} & \multicolumn{1}{c|}{\textbf{FP}} & 164 & 119 & 117 & \multicolumn{1}{c|}{} & \multicolumn{1}{c|}{\textbf{FP}} & 156 & 211 & 295 \\
\multicolumn{1}{|c|}{} & \multicolumn{1}{c|}{\textbf{FN}} & 101 & 102 & 99 & \multicolumn{1}{c|}{} & \multicolumn{1}{c|}{\textbf{FN}} & 168 & 167 & 168 \\ \hline
\multicolumn{1}{|c|}{\multirow{3}{*}{\textbf{Ent:AttrE}}} & \multicolumn{1}{c|}{\textbf{TP}} & 0 & 0 & 0 & \multicolumn{1}{c|}{\multirow{3}{*}{\textbf{Ent:AttrE}}} & \multicolumn{1}{c|}{\textbf{TP}} & 0 & 0 & 0 \\
\multicolumn{1}{|c|}{} & \multicolumn{1}{c|}{\textbf{FP}} & 5 & 4 & 5 & \multicolumn{1}{c|}{} & \multicolumn{1}{c|}{\textbf{FP}} & 11 & 15 & 19 \\
\multicolumn{1}{|c|}{} & \multicolumn{1}{c|}{\textbf{FN}} & 6 & 6 & 6 & \multicolumn{1}{c|}{} & \multicolumn{1}{c|}{\textbf{FN}} & 7 & 7 & 7 \\
\hline
\end{tabular}
}
\caption{Performance (FERRANTI: content-based categories, TP, FP, FN) of FactPegasus on SAMSum and DialogSum.}
\label{tab:factpegasus_samsum_dialogsum_tp_content}
\end{table*}

\begin{table*}[]
\resizebox{\linewidth}{!}{
\begin{tabular}{|crrr|crrr|}
\hline
\multicolumn{4}{|c|}{\textbf{SAMSum}} & \multicolumn{4}{c|}{\textbf{DialogSum}} \\ \hline
\multicolumn{1}{|c|}{Pre-trained Models} & \multicolumn{1}{c}{BART} & \multicolumn{1}{c}{BERT} & \multicolumn{1}{c|}{RoBERTa} & \multicolumn{1}{c|}{Pre-trained Models} & \multicolumn{1}{c}{BART} & \multicolumn{1}{c}{BERT} & \multicolumn{1}{c|}{RoBERTa} \\ \hline
\multicolumn{1}{|c|}{\textbf{FactCC}} & 0.2325 & 0.2290 & \textbf{0.2348} & \multicolumn{1}{c|}{\textbf{FactCC}} & 0.1273 & \textbf{0.1281} & \textbf{0.1281} \\ \hline
\multicolumn{1}{|c|}{\textbf{DAE}} & 0.1750 & \textbf{0.1756} & 0.1733 & \multicolumn{1}{c|}{\textbf{DAE}} & 0.0867 & \textbf{0.0892} & \textbf{0.0892} \\ \hline
\multicolumn{1}{|c|}{\textbf{QuestEval}} & 0.4793 & 0.4794 & \textbf{0.4807} & \multicolumn{1}{c|}{\textbf{QuestEval}} & 0.3793 & \textbf{0.3802} & 0.3799 \\ \hline
\multicolumn{1}{|c|}{\textbf{BARTScore}} & \textbf{-2.7799} & -2.7888 & -2.7879 & \multicolumn{1}{c|}{\textbf{BARTScore}} & -2.9237 & -2.9175 & \textbf{-2.9111} \\ \hline
\end{tabular}
}
\caption{Performance (factuality metrics) of CCGS on SAMSum and DialogSum. }
\label{tab:ccgs_samsum_dialogsum_fact}
\end{table*}

\begin{table*}[]
\resizebox{\linewidth}{!}{
\begin{tabular}{|crrr|crrr|}
\hline
\multicolumn{4}{|c|}{\textbf{SAMSum}} & \multicolumn{4}{c|}{\textbf{DialogSum}} \\ \hline
\multicolumn{1}{|c|}{Spacy Models} & \multicolumn{1}{c}{sm} & \multicolumn{1}{c}{md} & \multicolumn{1}{c|}{lg} & \multicolumn{1}{c|}{Spacy Models} & \multicolumn{1}{c}{sm} & \multicolumn{1}{c}{md} & \multicolumn{1}{c|}{lg} \\ \hline
\multicolumn{1}{|c|}{\textbf{FactCC}} & \textbf{0.2380} & 0.2273 & 0.2277 & \multicolumn{1}{c|}{\textbf{FactCC}} & \textbf{0.2282} & 0.2195 & 0.2278 \\
\multicolumn{1}{|c|}{\textbf{DAE}} & 0.1550 & \textbf{0.1625} & 0.1604 & \multicolumn{1}{c|}{\textbf{DAE}} & \textbf{0.1042} & 0.0819 & 0.0865 \\
\multicolumn{1}{|c|}{\textbf{QuestEval}} & 0.4671 & 0.4689 & \textbf{0.4726} & \multicolumn{1}{c|}{\textbf{QuestEval}} & 0.3822 & \textbf{0.3846} & 0.3844 \\
\multicolumn{1}{|c|}{\textbf{BARTScore}} & -2.9895 & \textbf{-2.9477} & -2.9650 & \multicolumn{1}{c|}{\textbf{BARTScore}} & \textbf{-3.0689} & -3.1092 & -3.1079 \\ \hline
\end{tabular}
}
\caption{Performance (factuality metrics) of FactPegasus on SAMSum and DialogSum. }
\label{tab:factpegasus_samsum_dialogsum_fact}
\end{table*}

\begin{table*}[]
\resizebox{\linewidth}{!}{
\begin{tabular}{|crrrr|crrrr|}
\hline
\multicolumn{5}{|c|}{\textbf{SAMSum}} & \multicolumn{5}{c|}{\textbf{DialogSum}} \\ \hline
\multicolumn{5}{|c|}{BART as the pre-trained model} & \multicolumn{5}{c|}{BART as the pre-trained model} \\ \hline
\multicolumn{1}{|l|}{} & \multicolumn{2}{c|}{\textbf{Pseudo}} & \multicolumn{2}{c|}{\textbf{Pseudo+Real}} & \multicolumn{1}{c|}{} & \multicolumn{2}{c|}{\textbf{Pseudo}} & \multicolumn{2}{c|}{\textbf{Pseudo+Real}} \\ \hline
\multicolumn{1}{|l|}{\textbf{corpus}} & \multicolumn{1}{c}{\textbf{SAMSum}} & \multicolumn{1}{c|}{\textbf{Mix}} & \multicolumn{1}{c}{\textbf{SAMSum}} & \multicolumn{1}{c|}{\textbf{Mix}} & \multicolumn{1}{c|}{\textbf{corpus}} & \multicolumn{1}{c}{\textbf{DialogSum}} & \multicolumn{1}{c|}{\textbf{Mix}} & \multicolumn{1}{c}{\textbf{DialogSum}} & \multicolumn{1}{c|}{\textbf{Mix}} \\ \hline
\multicolumn{1}{|c|}{\textbf{M}} & 0.00 & \multicolumn{1}{r|}{0.00} & 0.00 & 0.00 & \multicolumn{1}{c|}{\textbf{M}} & 5.49 & \multicolumn{1}{r|}{\textbf{5.75}} & 0.00 & 0.00 \\
\multicolumn{1}{|c|}{\textbf{R}} & 4.26 & \multicolumn{1}{r|}{\textbf{14.95}} & \textbf{15.00} & 11.63 & \multicolumn{1}{c|}{\textbf{R}} & 3.48 & \multicolumn{1}{r|}{\textbf{3.80}} & 1.74 & \textbf{2.51} \\
\multicolumn{1}{|c|}{\textbf{U}} & \textbf{7.04} & \multicolumn{1}{r|}{5.26} & \textbf{13.66} & 7.41 & \multicolumn{1}{c|}{\textbf{U}} & \textbf{12.05} & \multicolumn{1}{r|}{3.50} & 4.57 & \textbf{7.69} \\
\multicolumn{1}{|c|}{\textbf{Total}} & 4.15 & \multicolumn{1}{r|}{\textbf{11.58}} & \textbf{13.01} & 9.36 & \multicolumn{1}{c|}{\textbf{Total}} & \textbf{4.24} & \multicolumn{1}{r|}{3.91} & 2.31 & \textbf{3.47} \\
\hline
\end{tabular}
}
\caption{Performance (FERRANTI: form-based categories) of different pseudo-data corpus on SAMSum and DialogSum.  \textbf{Mix} means to mix the pseudo data constructed from SAMSum and DialogSum together. The values are all $\rm{F_{0.5}}$ scores. The better of the two pseudo-data corpus results is bolded. The data augmentation approach is set to \textbf{rule}.}
\label{tab:corpus_samsum_dialogsum_form}
\end{table*}

\end{document}